\documentclass{article}

\pdfoutput=1

\usepackage{microtype}
\usepackage{graphicx}
\usepackage{subfig}
\usepackage{booktabs} % for professional tables
\usepackage{mathtools}
\usepackage{hyperref}

\usepackage{algorithm}
\usepackage{algorithmic}

\usepackage{authblk}
\usepackage{fullpage}
\usepackage{Definitions}
\usepackage[round]{natbib}

\usepackage{multirow}

\title{Improving Expert Predictions with Conformal Prediction}

\author[1]{Eleni Straitouri}
\author[2]{Lequn Wang}
\author[1]{Nastaran Okati}
\author[1]{Manuel Gomez Rodriguez}

\affil[1]{Max Planck Institute for Software Systems, \{estraitouri, nastaran, manuelgr\}@mpi-sws.org}
\affil[2]{Cornell University, lw633@cornell.edu}

\date{}

\begin{document}

\maketitle

\begin{abstract}
Automated decision support systems promise to help human experts solve multiclass 
classification tasks more effi\-cient\-ly and accurately.
However, existing systems ty\-pi\-cally require experts to understand when to cede 
agency to the system or when to exer\-cise their own agency.
Otherwise, the experts may be better off solving the classification tasks on their 
own.
In this work, we develop an automated decision support system that, by design, does 
not require experts to understand when to trust the system to improve performance.
Rather than providing (single) label predictions and letting experts decide when to trust these predictions, our system provides sets of label predictions constructed using conformal 
prediction---prediction sets---and forcefully asks experts to predict labels from these 
sets.
By using conformal prediction, our system can precisely trade-off the probability 
that the true label is not in the prediction set, which determines how frequently
our system will mislead the experts, and the size of the prediction set, 
which determines the difficulty of the classification task the experts need 
to solve using our system.
% consider an 
% automated decision support system that, for each data sample, 
% \edit{simuesplifies the prediction task by using}
% uses a classifier to re\-commend a subset of labels% 
% \delete{to a human expert.}
% \edit{, from which a human expert has to predict the true label.
% 
% The smaller the re\-commended subset, the easier the task, though the higher the risk 
% of excluding the true label, thus misleading the human expert.
% We show that we can control the above trade-off, } 
%
% by looking at the design of such a system from the perspective of 
% conformal prediction,
%
% \edit{with which }%
% Alternative 1
% we can ensure that the probability that the recommended subset 
% of labels contains the true label, 
% \edit{\ie, does not mislead the expert,}
% matches almost exactly a target pro\-ba\-bi\-li\-ty 
% value with high probability.
% Alternative 2
% \edit{
% we can control almost exactly with high probability how often the recommended subset will (not) include the true label, \ie,  how often the system will simplify the task (mislead the expert). 
% 
% Then, we develop an efficient and near-optimal search method to trade-off simplifying the task and misleading the expert so that the expert benefits the most from using our system.
% }
%
In addi\-tion, we develop an efficient and near-optimal search method to find the 
conformal predictor under which the experts benefit the most from using our system.
% target probability value under which the expert benefits the most from using our system.
% 
Simulation experiments using synthetic and real expert predictions demonstrate that our system may help experts make more accurate predictions and is robust to the accuracy of the classifier the conformal predictor relies on. 
\end{abstract}

% \vspace{-2mm}
\section{Introduction}\label{sec:intro}
% \vspace{-2mm}
In recent years, there has been an increasing interest in de\-ve\-lo\-ping automated decision 
support systems to help human experts solve tasks in a wide range of critical domains, from medicine~\citep{jiao2020deep} and drug dis\-co\-ve\-ry~\citep{liu2021evaluating} to candidate screening~\citep{wang2022improving} and criminal justice~\citep{grgic2019human}, to name a few.
Among them, one of the main focuses has been multiclass classi\-fi\-ca\-tion tasks, where a decision support system uses a classifier to make label predictions and the experts decide when to follow the predictions made by the classifier~\citep{bansal2019beyond, lubars2019ask, bordt2020humans}.

However, these systems typically require human experts to un\-ders\-tand when to trust 
a prediction made by the classi\-fier. Otherwise, the experts may be better off solving the 
classification tasks on their own~\citep{suresh2020misplaced}.
This follows from the fact that, in general, the accuracy of a classifier differs 
across data samples~\citep{raghu2019algorithmic}.
In this context, several recent studies have analyzed how factors such as model confidence, 
model explanations and overall model calibration modulate trust~\citep{papenmeier2019model, wang2021explanations, vodrahalli2022uncalibrated}.
Unfortunately, it is not yet clear how to make sure that the experts do not develop a misplaced trust 
that decreases their performance~\citep{yin2019understanding, nourani2020role, zhang2020effect}.
In this work, we develop a decision support\- system for multiclass classification tasks that, by design, does not require experts to understand when to trust the system to improve their performance.
%
% \vspace{-1mm}

\xhdr{Our contributions}
%
% We develop a decision support system that, 
% Rather than providing a single label prediction for each data sample, 
% \edit{simplify the prediction task by using}
For each data sample, our decision support system provides a set of label 
predictions---a prediction set---and forcefully asks human experts to predict 
a label from this set\footnote{There are many systems used everyday by experts 
(\eg, pilots flying a plane) that, under normal operation, restrict their choices. 
This does not mean that, in extreme circumstances, the expert should not have 
the ability to essentially switch off the system.}.
We view this type of decision support system as more natural since, given a set of 
alternatives, experts tend to narrow down their options to a subset of them 
before making their final decision~\citep{wright1977phased,beach93, BENAKIVA19959}.
In a way, our support system helps experts by au\-to\-ma\-ti\-cally narrowing down 
their options for them, decreasing their cognitive load and allowing them to focus their
attention where it is most needed. 
This could be particularly useful when the task is tedious or requires domain knowledge since it is difficult to outsource the task, and domain experts are often a scarce resource.
In the context of clinical text annotation\footnote{Clinical text annotation is a task where medical experts aim to identify clinical concepts in medical notes and map them to labels in a large ontology.}, a recent empirical study has concluded that, in terms of the overall accuracy, it may be more beneficial to recommend a subset of options than a single option~\citep{sontag21}. 

By using the theory of conformal prediction~\citep{VovkBook,angelopoulos2021gentle} 
to construct the above prediction set, 
our system can precisely control the trade-off between the probability that the true label is not in the prediction set, which determines how frequently our system will mislead an expert, and the size of the prediction set, which determines the difficulty
of the classification task the expert needs to solve using our system.
In this context, note that, if our system would not forcefully ask the expert to 
predict a label from the prediction set, 
it would not be able to have this level of control and good performance would depend 
on the expert developing a good sense on when to predict a label from the prediction 
set and when to predict a label from outside the set, as noted by~\citet{sontag21}.
In addition, given an estimator of the expert'{}s success probability for any of the possible prediction sets, we develop an efficient and near-optimal search method to find the conformal predictor under which the expert is guaranteed to achieve the greatest 
accuracy with high probability.
% 
% \edit{to discover the optimal trade-off between simplifying the prediction task and the 
% risk of misleading the expert }%
% given an estimator of the expert'{}s success probability for any of the recommended 
% subsets, we develop an efficient and near-optimal search method to find the target 
% probability value under which the expert is guaranteed to achieve the greatest accuracy 
% with high probability.
%
In this context, we also propose a practical method to obtain such an estimator using the confusion matrix of the expert predictions in the ori\-gi\-nal classification task and a given discrete choice model. 

Finally, we perform simulation experiments using synthetic and real expert predictions
on several multiclass classification tasks. 
The results demonstrate that our decision support system is robust to both the accuracy of the classifier and the estimator of the expert's success probability it relies on---the competitive advantage it provides improves with their accuracy, and the human experts do not decrease their performance by using the system even if the classifier or the estimator are very inaccurate. 
Additionally, the results also show that, even if the classifiers that our system relies on have high accuracy, an expert using our system may achieve significantly higher accuracy
than the classifiers on their own---in our experiments with real data, the relative reduction in misclassification probability is over $72$\%.
Finally, by using our system, our results suggest that the expert would reduce their misclassification pro\-ba\-bi\-lity by $\sim$$80$\%\footnote{
% To facilitate research in this area, 
% we will release an open-source implementation of our system with the final version of the paper. We have submitted an anonymized version of the code with the supplementary material.}.
An open-source implementation of our system is avai\-la\-ble at \href{https://github.com/Networks-Learning/improve-expert-predictions-conformal-prediction}{https://github.com/Networks-Learning/improve-expert-predictions-conformal-prediction}.}.

\xhdr{Further related work}
Our work builds upon further related work on distribution-free uncertainty quantification, reliable classification and learning
under algorithmic triage.
% NOTE: The term 'reliable classification' does not seem broadly used in 
% the literature, however it is true that the motivation and goal in all the pointers
% given by the reviewer as well as works on set-valued classifiers are indeed
% to make the classifier predictions more reliable. However, one could tell 
% that also conformal prediction can be considered as 'reliable classification'.
% The pointers seem to relate to the line of works of nondeterministic classification
% in which the output of a classifier may include more label values in case 
% of uncertainty or even the option to 'reject' (defer) the prediction.
% In contrast to our work, this line of works does not assume that a human
% expert makes the final prediction using the set-valued prediction,
% but rather defines some notion of utility to evaluate the set-valued
% predictions and specifically train the classifiers to maximize this utility.
% 
% (The citation on multilabel classification seems less relevant.)

There exist three fundamental notions of distribution-free uncertainty quantification in the literature: calibration, confidence intervals, and prediction sets~\citep{VovkBook, balasubramanian2014conformal, gupta2020distribution, angelopoulos2021gentle}.
Our work is most closely related to the rapidly increasing literature on prediction sets~\citep{romano2019conformalized,romano2020classification,angelopoulos2021uncertainty,podkopaev2021distribution}, however, to the best of our knowledge, prediction sets have not been optimized to serve automated decision support systems such as ours. 
In this context, we acknowledge that~\citet{babbar2022utility} have also very recently proposed using prediction sets in decision support systems.
% In this context, we acknowledge that~\citet{babbar2022utility} have also proposed using prediction sets in decision support systems, however, this work is contemporary to ours and has been carried out independently.
%
% Moreover,
However, in contrast to our work, for each data sample, 
% \edit{the system of~\citet{babbar2022utility}  recommends a subset of alternatives, \ie, a subset of labels, to the expert, but the expert is allowed to choose any alternative from the universe of alternatives, \ie, the label set, not only those in the subset of alternatives.
% Moreover,~\citet{babbar2022utility} do not optimize the probability that the true label belongs to the subset.
% }
they allow the expert to predict label values outside the recommended subset, \ie, to predict any alternative from the entire universe of alternatives, and do not optimize the probability that the true label belongs to the subset.
% (to improve expert predictions with high probability).
%
As a result, their method is not directly comparable to ours\footnote{In our simulation experiments, we estimate the performance achieved by an expert using our system via a model-based estimator of the expert'{}s success probability. Therefore, to compare our system with the system by~\citet{babbar2022utility}, we would need to model the expert'{}s success probability whenever the expert can predict any label given a prediction set, a problem for which discrete choice theory provides little guidance.}.
% \edit{
% One would need to resort to costly interventional experiments to compare the methods, 
% as little  guidance is provided by 
% the discrete choice theory and~\citet{babbar2022utility} about modeling the expert 
% success probability, when an expert can predict any alternative from the entire 
% universe of alternatives, given a recommended subset of alternatives. 
% }

There is an extensive line of work on reliable or cautious classification~\citep{del2009learning,liu2014credal, yang2017cautious,mortier2021efficient,ma2021partial,nguyen2021multilabel}.
Reliable classification aims to develop models that can provide set-valued predictions to account for the prediction uncertainty of a classifier. 
However, in this line of work, there are no human experts who make the final predictions given the set-valued predictions, in contrast with our 
work.
Moreover, the set-valued predictions typically lack distribution-free~guarantees.
% e.g. belief functions, imprecise probabilities
% that lack distribution-free guarantees,
% contrary to prediction sets in the context of distribution-free uncertainty quantification.
% 
% In addition, prior work on cautious classification, in contrast to our work, does not consider that human experts make the final prediction using the set-valued predictions, 
% but rather directly evaluate the set-valued predictions themselves using various performance measures. 
%}

Learning under algorithmic triage seeks the development of machine learning models that operate 
under different automation levels---models that make decisions for a given fraction of instances 
and leave the remaining ones to human experts~\citep{raghu2019algorithmic, mozannar2020consistent, de2020regression, de2021classification,  okati2021differentiable}. 
This line of work has predominantly focused on supervised learning settings with a few very
recent notable exceptions~\citep{straitouri2021reinforcement, meresht2020learning}.
However, in this line of work, each sample is either predicted by the model or by the human expert. 
In contrast, in our work, the model helps the human predict each sample.
% \edit{
That being said, it is worth noting that there may be classifiers, data distributions and conformal scores under which the optimal conformal predictor and the optimal triage policy coincide, \ie, the optimal conformal predictor does recommend a single label or the entire label set of labels.
% our system could capture the behavior of the optimal triage policy if the model, the 
% data distribution, and the conformal score were such that the optimal conformal 
% predictor would only recommend either a single label, 
% ---equivalent to the model predicting the label of the sample---or the entire label set 
% of labels--equivalent to the human expert predicting alone the label of the sample.
% }

% \vspace{-1mm}
\section{Problem Formulation}
\label{sec:problem}
% \vspace{-1mm}
\begin{figure*}[t]
\centering
% \vspace{-3mm}
\includegraphics[width=.8\textwidth]{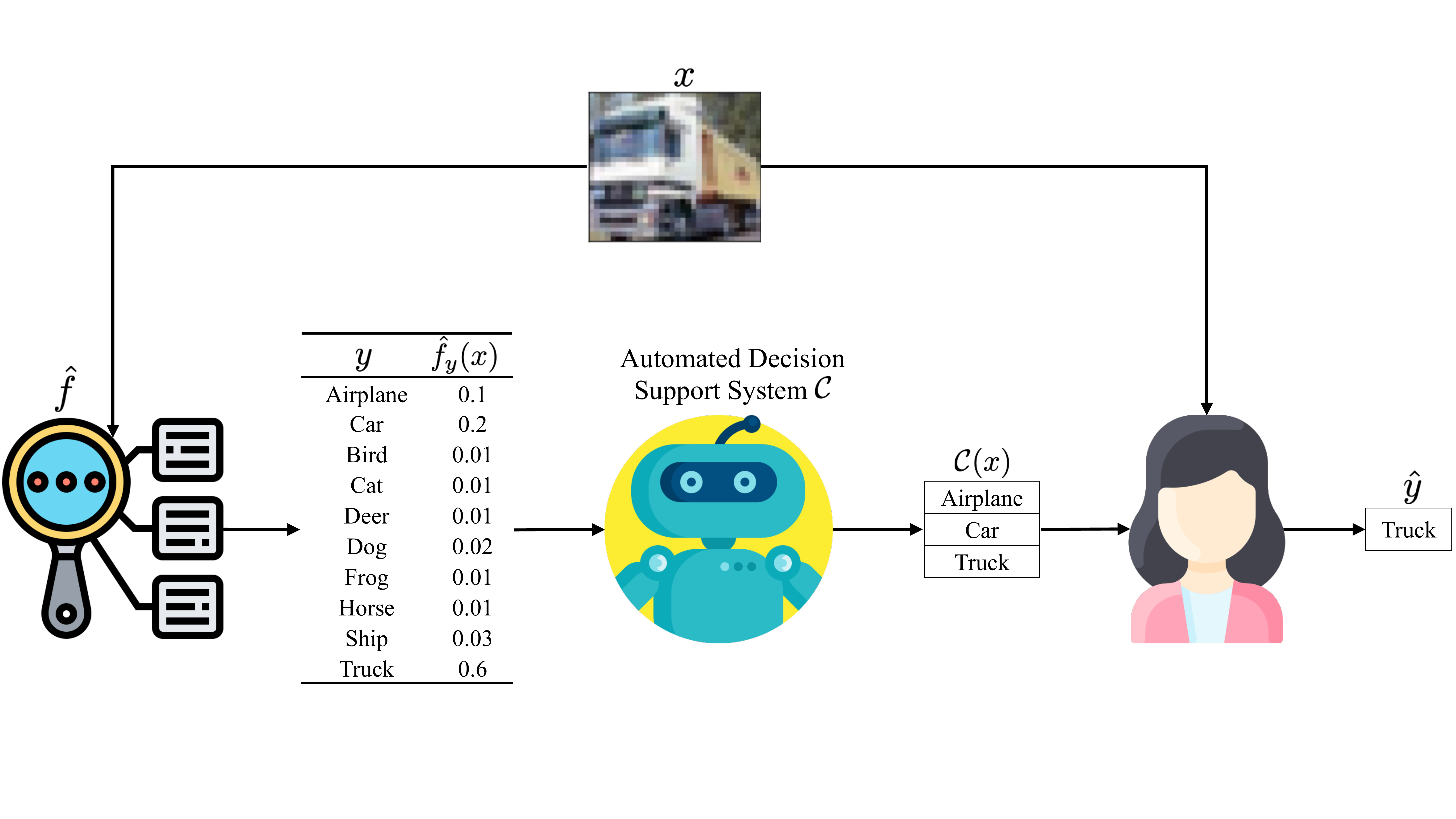}
\vspace{-7mm}
\caption{Our automated decision support system $\Ccal$. 
Given a sample with a feature vector $x$, our system $\Ccal$ narrows down the set of potential labels $y \in \Ycal$ to a subset of them $\Ccal(x)$ using the scores $\hat{f}_{y}(x)$ provided by a classifier $\hat{f}$ for each class $y$. 
The human expert receives the recommended subset $\Ccal(x)$, together with the sample, and predicts a label $\hat{y}$ from $\Ccal(x)$ according to a policy $\pi(x, \Ccal(x))$.
}
\label{fig:illustrative}
\end{figure*}
We consider a multiclass classification task where a human expert observes 
a feature vector\footnote{We denote random variables with capital letters and realizations of random variables with lower case letters. } $x \in \Xcal$, with $x \sim P(X)$, and needs to predict a label $y \in \Ycal = \{1,
\ldots ,n \}$, with $y \sim P(Y \given X)$.
Then, our goal is to design an automated decision support system $\Ccal:\Xcal \rightarrow 2^{\Ycal}$ that, given a feature vector $x \in \Xcal$, helps the expert by automatically narrowing down the set of potential labels to a subset of them $\Ccal(x) \subseteq \Ycal$ using a trained classifier 
% \delete{$\hat{f}(x) \in [0, 1]^{n}$}
% \edit{
$\hat{f}:\Xcal \rightarrow [0,1]^{|\Ycal|}$
%} 
that outputs scores for each class (\eg, softmax scores)\footnote{The assumption that $\hat{f}(x) \in [0,1]^{n}$ is without loss of generality.}. The higher the score $\hat {f}_y(x)$, the more the classifier believes the true label $Y = y$.
Here, we assume that, for each $x \sim P(X)$, the expert predicts a label $\hat Y$ among those in the subset $\Ccal(x)$ according to an unknown po\-licy $\pi(x, \Ccal(x))$.
More formally, $\hat Y \sim \pi(x, \Ccal(x))$, where $\pi : \Xcal \times 2^{\Ycal} \rightarrow \Delta(\Ycal)$ and $\Delta(\Ycal)$ denotes the probability simplex over the set of labels $\Ycal$, and $\pi_{y}(x, \Ccal(x)) = 0$ if $y \notin \Ccal(x)$.
Refer to Figure~\ref{fig:illustrative} for an illustration of the automated decision support system we consider.

Ideally, we would like that, by design, 
the expert can only benefit from using the automated decision support system $\Ccal$,
\ie,
\begin{equation} \label{eq:accuracy-always-better}
    \PP[\hat{Y} = Y \,;\, \Ccal] \geq \PP[\hat{Y} = Y \,;\, \Ycal],
\end{equation}
where $\PP[\hat{Y} = Y \,;\, \Ccal]$ denotes the expert'{}s success probability
if, for each $x \sim P(X)$, the human expert predicts a label $\hat Y$ among those in the subset $\Ccal(x)$.
However, not all automated decision support systems ful\-filling the above requirement will be equally useful---some will help experts increase their success probability more than others.
For exam\-ple, a system that always recommends $\Ccal(x) = \Ycal$ for all $x \in \Xcal$ satisfies Eq.~\ref{eq:accuracy-always-better}. However, it is useless to the experts. 
Therefore, among those systems satisfying Eq.~\ref{eq:accuracy-always-better}, we would % ideally
like to find the system $\Ccal^{*}$ that helps the experts achieve the highest success probability\footnote{Note that maximizing the expert'{}s success probability $\PP[\hat{Y} = Y \,;\, \Ccal]$ is equi\-va\-lent to minimizing the expected $0$-$1$ loss $\EE[\II(\hat{Y} \neq Y) \,;\, \Ccal]$. Considering other types of losses is left as an interesting avenue for future work.}, \ie, 
\begin{equation} \label{eq:optimal-accuracy}
    \Ccal^{*} = \argmax_{\Ccal} \, \PP[\hat{Y} = Y \,;\, \Ccal].
\end{equation}
To address the design of such a system, we will look at the problem from the perspective of conformal prediction~\citep{VovkBook, angelopoulos2021gentle}.

% \vspace{-1mm}
\section{Subset Selection using Conformal Prediction}
\label{sec:conformal-prediction}
% \vspace{-1mm}
In general, if the trained classifier $\hat{f}$ we use to build $\Ccal(X)$ is not perfect, the true 
label $Y$ may or may not be included in $\Ccal(X)$. 
In what follows, we will construct the subsets $\Ccal(X)$ using the theory of conformal prediction. 
This will allow our system to be robust to the accuracy of the classifier $\hat{f}$ it uses---the probability 
$\PP[ Y \in \Ccal(X)]$ that the true label $Y$ belongs to the subset $\Ccal(X) = \Ccal_{\alpha}(X)$ will 
match almost exactly a given target probability $1-\alpha$ with high probability, without making 
any distributional assumptions about the data distribution $P(X)P(Y \given X)$ nor the classifier $\hat{f}$.

Let $\Dcal_{\text{cal}} = \{(x_i,y_i)\}_{i=1}^{m}$ be a calibration set, where $(x_i, y_i) \sim P(X) P(Y \given X)$,
$s(x_i, y_i) = 1 - \hat{f}_{y_i}(x_i)$ be the \emph{conformal score}\footnote{In general, the conformal score $s(x,y)$ can be any function of $x$ and $y$ measuring the \emph{similarity} between samples.}
(\ie, if the classifier is catastrophically wrong, the conformal score will be close to one),
and $\hat{q}_{\alpha}$ be the $\frac{\lceil(m + 1)(1 - \alpha)\rceil}{m}$ empirical quantile of the conformal scores $s(x_1, y_1), \ldots, s(x_m, y_m)$.
Then, if we construct the subsets $\Ccal_{\alpha}(X)$ for new data samples as follows:
\begin{equation} \label{eq:prediction-set}
\Ccal_{\alpha}(X) = \{y \given s(X, y) \leq \hat{q}_{\alpha} \},
\end{equation}
we have that the probability that the true label $Y$ belongs to the subset $\Ccal_{\alpha}(X)$ conditionally on the calibration set $\Dcal_{\text{cal}}$ is almost exactly $1-\alpha$ with high probability as long as the size $m$ of the calibration set is sufficiently large.
Specifically, we first note that the coverage probability is a random quantity\footnote{% \edit{
The randomness comes from the randomness of the calibration set sampling.}
% } 
whose distribution is given by the following proposition (refer to Appendix~A.5 in~\citet{hulsman2022distribution} for the proof):
\begin{proposition}\label{prop:coverage-beta}
    For a decision support system $\Ccal_{\alpha}$ that constructs the subsets $\Ccal_{\alpha}(X)$ using Eq.~\ref{eq:prediction-set}, it holds that
    \begin{equation}\label{eq:coverage-beta}
        \PP[Y \in \Ccal_{\alpha}(X) \given \Dcal_{\text{cal}}] \sim \text{Beta}(\lceil(m+1)(1 - \alpha)\rceil, \lfloor(m+1)\alpha\rfloor)
    \end{equation}
    as long as the conformal scores $s(X_i, Y_i)$ for all $(X_i, Y_i) \in \Dcal_{\text{cal}}$ are almost surely distinct. 
    % the probability is over the randomness in the sample that the system $\Ccal_{\alpha}$ helps predicting.
% with equality holding for almost surely distinct conformal scores.
\end{proposition}
As an immediate consequence of Proposition~\ref{prop:coverage-beta}, using the definition of the beta distribution, we have that
\begin{align*}
% = \frac{\lceil(1 - \alpha)(m+1)\rceil}{\lceil(1 - \alpha)(m+1)\rceil + \lfloor\alpha(m+1)\rfloor} \approx 1-\alpha
1-\alpha &\leq \EE\left[ \PP[Y \in \Ccal_{\alpha}(X) \given \Dcal_{\text{cal}}] \right] = 1 - \frac{\lfloor (m+1)\alpha \rfloor}{m+1}\\& \leq 1 - \alpha + \frac{1}{m+1}.
\end{align*}
%
% It follows directly that the mean value of the above beta distribution is $\frac{\lceil(1 - \alpha)(m+1)\rceil}{\lceil(1 - \alpha)(m+1)\rceil + 
% \lfloor\alpha(m+1)\rfloor} \approx 1 - \alpha$. 
%
Moreover, given a target probability $1 - \alpha$ and tolerance values $\delta, \epsilon \in (0, 1)$, we can compute the minimum size $m$ of the calibration set $\Dcal_{\text{cal}}$
such that $\Ccal_{\alpha}$ enjoys Probably Approximately Correct (PAC) coverage guarantees, \ie, with probability $1-\delta$, it holds that~\cite{angelopoulos2021gentle} 
\begin{equation*}
1 - \alpha - \epsilon \leq \PP[Y \in \Ccal_{\alpha}(X) \given \Dcal_{\text{cal}}] \leq 1 - \alpha + \epsilon.
\end{equation*}
%
% $\PP[Y \in \Ccal_{\alpha}(X) \given \Dcal_{\text{cal}}]$ lies in $1 - \alpha \pm \epsilon$ with probability $1 - \delta$, for some 
% $\epsilon,\delta \in (0,1)$. For a given $\alpha$ value, a sufficiently large $m$ would allow $\epsilon,\delta$ to 
% be small, \ie, $\PP[Y \in \Ccal_{\alpha}(X) \given \Dcal_{\text{cal}}]$ to be almost exactly $1-\alpha$ with high probability~\citep{angelopoulos2021gentle}. 
% This result implies that for a given $\alpha$ value, the system $\Ccal_{\alpha}$ can have Probably Approximately Correct (PAC)
% coverage guarantees given a fixed calibration set $\Dcal_\text{cal}$.
% From the above follows directly one could choose $m$ so that  
% $\PP[Y \in \Ccal_{\alpha}(X) \given \Dcal_{\text{cal}}]$ will be close to $1 - \alpha$ with high 
% probability~\citep{angelopoulos2021gentle}.
% and a sufficiently large $m$,
% }
%
% \delete{
% \begin{theorem} %\label{thm:standard-coverage}
% For an automated decision support system $\Ccal_{\alpha}$ that constructs the subsets 
% $\Ccal_{\alpha}(X)$ using Eq.~\ref{eq:prediction-set}, it holds that
%
% \begin{equation*}
% \label{eq:coverage}
% 1 - \alpha \leq \PP[Y \in \Ccal_{\alpha}(X)] \leq 1 - \alpha + \frac{1}{m+1},
% \end{equation*}
%
% where the probability is over the randomness in the sample it helps predicting and the calibration set used to compute the empirical quantile $\hat{q}_{\alpha}$.
%
% \end{theorem}
% }
%
% , given a fixed calibration set $\Dcal_{\text{cal}}$,
%
While the above coverage guarantee is valid for any choice of $\alpha$ value, we would like to emphasize that there may be some $\alpha$ values that will lead to larger gains 
in terms of success probability $\PP[\hat Y = Y \,;\, \Ccal_{\alpha}]$ than others. 
Therefore, in what follows, our goal is to find the optimal $\alpha^{*}$ that maximizes the expert'{}s success probability given a calibration set $\Dcal_{\text{cal}}$.

% \edit{
\xhdr{Remark}
Most of the literature on conformal prediction focuses on the following conformal calibration guarantee (refer to Appendix~D 
in~\citet{angelopoulos2021gentle} for the proof):
\begin{theorem} \label{thm:standard-coverage}
    For an automated decision support system $\Ccal_{\alpha}$ that constructs the subsets 
    $\Ccal_{\alpha}(X)$ using Eq.~\ref{eq:prediction-set}, it holds that
    \begin{equation*}
    % \label{eq:coverage}
     1 - \alpha \leq \PP[Y \in \Ccal_{\alpha}(X)] \leq 1 - \alpha + \frac{1}{m+1},
    \end{equation*}
    where the probability is over the randomness in the sample it helps predicting and the calibration set used to compute the empirical quantile $\hat{q}_{\alpha}$.
\end{theorem}
However, to afford the above marginal guarantee in our work, we would be unable to optimize the $\alpha$ value to maximize the expert' {}s success probability given a calibration
set $\Dcal_{\text{cal}}$. This is because the guarantee requires that $\alpha$  and $\Dcal_{\text{cal}}$ are independent.
That being said, in our experiments, we have empirically found that the optimal $\alpha^{*}$ does not vary significantly across calibration sets, as shown in Appendix~\ref{app:conf-guarantees}.

% \vspace{-1mm}
\section{Optimizing Across Conformal Predictors}
\label{sec:optimizing-conformal-prediction}
% \vspace{-1mm}
%
% fixed
We start by realizing that, given a calibration set $\Dcal_{\text{cal}} = \{(x_i, y_i)\}_{i=1}^{m}$, there only exist $m$ different conformal predictors. 
This is because the empirical quantile $\hat{q}_{\alpha}$, which the subsets $\Ccal_{\alpha}(x_i)$
depend on, can only take $m$ different values.
As a result, to find the optimal conformal predictor that maximizes the expert'{}s success
probability, we need to solve the following maximization problem:
\begin{equation} \label{eq:optimal-alpha}
     \alpha^{*} = \argmax_{\alpha \in \Acal} \, \PP[\hat Y = Y ; \Ccal_{\alpha}],
\end{equation}
where $\Acal = \{\alpha_i\}_{i \in [m]}$, with $\alpha_i = 1 - i/(m+1)$, and
the probability is only over the randomness in the samples the system helps predicting.

However, to find a near optimal solution $\hat{\alpha}$ to the above problem, 
we need to estimate the expert'{}s success pro\-ba\-bi\-lity $\PP[\hat Y = Y \,;\, \Ccal_{\alpha}]$. 
Assume\- for now that, for each $\alpha \in \Acal$,
we have access to an estimator $\hat{\mu}_{\alpha}$ of the expert'{}s success probability such that, for any $\delta \in (0, 1)$, with probability at least $1 - \delta$, it holds that $|\hat{\mu}_{\alpha} - \PP[\hat Y = Y \,;\, \Ccal_{\alpha}]| \leq \epsilon_{\alpha, \delta}$.
Then, we can use the following proposition to find a near-optimal solution $\hat{\alpha}$ to Eq.~\ref{eq:optimal-alpha} with high probability:
\begin{proposition}
\label{prop:improve-error-practice}
For any $\delta \in (0, 1)$, consider an automated decision support system $\Ccal_{\hat{\alpha}}$ with
\begin{equation} \label{eq:near-optimal-alpha}
    \hat{\alpha} = \argmax_{\alpha \in \Acal} \, \hat{\mu}_{\alpha} - \epsilon_{\alpha, \delta / m}.
\end{equation}
With probability at least $1-\delta$, it holds that
$\PP[\hat Y = Y \,;\,  \Ccal_{\hat{\alpha}}]  \geq \PP[\hat Y = Y \,;\,  \Ccal_{\alpha}] 
- 2 \epsilon_{\alpha, \delta / m}$ $\forall \alpha \in \Acal$ simultaneously.
\end{proposition}
More specifically, the above result directly implies that for any $\delta \in (0, 1)$, 
with pro\-ba\-bi\-lity at least $1-\delta$, it holds that:
\begin{align} \label{eq:near-optimal-distance}
    \PP[\hat Y = Y \,;\, \Ccal_{\alpha^{*}}] - \PP[\hat Y = Y \,;\,  \Ccal_{\hat{\alpha}}] \leq 2 \epsilon_{\alpha^{*}, \delta / m}. % \sqrt{\frac{(n - 1)^{2}\log{\frac{m}{\delta}}}{2k_{\alpha} n^2}}.
\end{align}
%
% QUESTION: Should we refer to both the marginal and the PAC coverage guarantees here to 
% make clear that the near-optimality of alpha does not depend on conformal prediction?
%
% To avoid misunderstandings, I think we should change "marginal guarantees" to "PAC coverage guarantees".
% I think that answers your question.
%
Here, note that the above guarantees do not make use of the PAC coverage guarantees afforded by conformal 
prediction---they hold for any parameterized set-value predictor.
\begin{algorithm}[t]
\begin{algorithmic}[1]
\STATE{{\bf Input:} $\hat{f}$, $\Dcal_{\textnormal{est}}$, $\Dcal_{\textnormal{cal}}$, $\delta$, $m$}
\STATE{{\bf Initialize:} $\Acal = \{\}, \hat{\alpha} \leftarrow 0, t \leftarrow 0$}
\vspace{1mm}
\FOR{$i \in {1,..,m}$}
\STATE{$\alpha \leftarrow 1 - \frac{i}{m+1}$}
\STATE{$\Acal \leftarrow \Acal \cup \{\alpha\}$}
\ENDFOR
\vspace{1mm}
\FOR{$\alpha \in \Acal$}
\STATE{\mbox{$\hat{\mu}_\alpha, \epsilon_{\alpha, \delta / m} \leftarrow \textsc{Estimate}(\alpha, \delta, \Dcal_{\textnormal{est}}, \Dcal_{\textnormal{cal}}, \hat{f})$}}
% \COMMENT{It uses Eqs.~\ref{eq:hat-mu} and~\ref{eq:epsilon}}
\IF{$t \leq \hat{\mu}_\alpha - \epsilon_{\alpha, \delta / m}$}
\STATE{$t \leftarrow \hat{\mu}_\alpha - \epsilon_{\alpha, \delta / m}$}
\STATE{$\hat{\alpha} \leftarrow \alpha$}
\ENDIF
\ENDFOR
\vspace{1mm}
\STATE{{\bf return} $\hat{\alpha}$}
\end{algorithmic}
\caption{Finding a near-optimal $\hat{\alpha}$}
\label{alg:near-optimal-alpha}
\end{algorithm}

In what follows, we propose a practical method to estimate the expert'{}s success 
probability $\PP[\hat Y = Y \,;\, \Ccal_{\alpha}]$ that builds upon the multinomial 
logit model (MNL), one of the most popular models in the vast literature on discrete 
choice models~\citep{heiss2016discrete}.
More specifically, given a sample $(x, y)$, we assume that the expert'{}s conditional
success probability for the subset $\Ccal_{\alpha}(x)$ is given by
\begin{equation}
    \label{eq:mnl-pred-prob}
    \PP[\hat Y = y \,;\, \Ccal_{\alpha} \given y \in \Ccal_{\alpha}(x)] = \frac{e^{ u_{yy} }}{\sum_{y' \in \Ccal_{\alpha}(x)} e^{ u_{yy'}}},
\end{equation}
where $u_{yy'}$ denotes the expert's preference for label value $y' \in \Ycal$ whenever 
the true label is $y$.
In the language of discrete choice models, one can view the true label $y$ as the \textit{context} in which the expert chooses among alternatives~\citep{tversky1993context}.
In Appendix~\ref{app:extra-expert-models}, we consider and experiment with a 
more expressive context that, in addition to the true label, distinguishes 
between diffe\-rent levels of \emph{difficulty} across data samples.
%
% note that the right-hand side of Eq.~\ref{eq:mnl-pred-prob} does depend on the 
% \textit{context}, \ie, the true label $y$, and also on the label values in the prediction 
% set $C_{\alpha}(x)$, but not on the specific value 
% of $x$.\footnote{\scriptsize In Appendix~\ref{app:extra-expert-models}, we experiment with 
% a more expressive model in which the \textit{context} also includes the ``difficulty'' of 
% each sample $x$.}}
% \delete{does not depend on the specific value 
% of $x$ but on the label values in the prediction set $C_{\alpha}(x)$.}

Further, to estimate the parameters $u_{yy'}$, we assume we have access to (an estimation of) the confusion matrix $\Cb$ for the expert predictions in the (original) multiclass classification task, similarly as in~\citet{kerrigan2021combining}, \ie, 
\begin{equation*}
\Cb = [C_{yy'}]_{y, y'\in \Ycal},\, \text{where} \, C_{yy'} = 
\PP[ \hat{Y} = y' \,;\, \Ycal \given Y = y ],
\end{equation*}
and naturally set $u_{yy'} = \log C_{yy'}$.
%
% In practice, we can compute a Monte-Carlo estimator $\hat{\mu}_{\alpha}$ 
% of the above conditional 
% success probability $\PP[\hat Y = y \,;\, \Ccal_{\alpha} \given y \in 
% \Ccal_{\alpha}(x)]$ using an 
% estimation set $\Dcal_{\text{est}} = \{ (x_i, y_i) \}_{i\in [m]}$
Then, we can compute a Monte-Carlo estimator $\hat{\mu}_{\alpha}$ 
of the expert'{}s success probability $\PP[\hat{Y} = Y\,;\,\Ccal_{\alpha}]$ using the above conditional success probability $\PP[\hat Y = y \,;\, \Ccal_{\alpha} \given y \in \Ccal_{\alpha}(x)]$ and an estimation set $\Dcal_{\text{est}} = \{ (x_i, y_i) \}_{i\in [m]}$\footnote{The number of samples in $\Dcal_{\text{cal}}$ and $\Dcal_{\text{est}}$ can differ. For simplicity, we assume both sets contain $m$ samples.}, \ie,
\begin{equation}
    \label{eq:hat-mu}
    \hat{\mu}_{\alpha} = \frac{1}{m} \sum_{i \in [m] \given y_i \in \Ccal_{\alpha}(x_i)} \PP[\hat Y = y_i \,;\, \Ccal_{\alpha} \given y_i \in \Ccal_{\alpha}(x_i)].
\end{equation}
%
% where $k_{\alpha} = \sum_{i \in [m]} \II\{ y_i \in \Ccal_{\alpha}(x_i) \}$. 
%
Finally, for each $\alpha \in \Acal$, using Hoeff\-ding'{}s inequality\footnote{By using Hoeffding'{}s
inequality, we derive a fairly conservative constant error bound for all $\alpha$ values, however, we have experimentally found that, even with a relatively small amount of estimation and calibration data, our algorithm identifies near-optimal 
$\hat \alpha$ values, % providing large gains 
% with respect to $\PP[\hat Y = Y; \Ycal]$ both in synthetic and real data, 
as shown in % Tables~\ref{table:synthetic-acc-10-labels} and~\ref{table:real-acc-labels} and 
% Appendix~\ref{app:alpha-values}.
Figure~\ref{fig:synthetic_error_alpha}.
% \ref{app:error_alpha}. 
%
That being said, one could use tighter concentration inequalities such as Hoeffding–Bentkus and Waudby-Smith–Ramdas~\citep{bates2021distributionfree}.}\textsuperscript{,}\footnote{% \edit{
We are applying Hoeffding'{}s inequality only on the randomness of the samples $(X_i, Y_i)$, which are independent and identically distributed.},
% , and we treat $\Ccal_{\alpha}$ and the conditional expert success probability as constants.}, 
we can conclude that, with probability at least $1-\delta$, it holds that (refer to Appendix~\ref{app:hoeffding}):
\begin{equation} \label{eq:epsilon}
|\hat{\mu}_{\alpha} - \PP[\hat Y = Y \,;\,  \Ccal_{\alpha}]| \leq  \sqrt{\frac{\log{\frac{1}{\delta}}}{2m}} \coloneqq \epsilon_{\alpha, \delta}.
\end{equation}
As a consequence, as $m \rightarrow \infty$, $\epsilon_{\alpha, \delta}$ converges to zero. 
This directly implies that the near-optimal $\hat{\alpha}$ converges to the true optimal
$\alpha^{*}$ and that, with probability at least $1 - \delta$, our system $\Ccal_{\alpha^{*}}$ satisfies 
Eq.~\ref{eq:accuracy-always-better} % and~\ref{eq:optimal-accuracy} %\edit{ for the predictors 
% corresponding to the fixed calibration set $\Dcal_{\text{cal}}$, }
asymptotically with respect to the number of samples $m$ in the estimation set.

Algorithm~\ref{alg:near-optimal-alpha} summarizes the overall search method, 
where the function $\textsc{Estimate}(\cdot)$ uses Eqs.~\ref{eq:hat-mu} and~\ref{eq:epsilon}.
The algorithm first builds $\Acal$ and then finds the near-optimal 
$\hat{\alpha}$ in $\Acal$.
To build $\Acal$, it needs $\Ocal(m)$ steps.
To find the near-optimal $\hat{\alpha}$, for each value $\alpha \in \Acal$ and each sample $(x_i, y_i) \in \Dcal_{\text{est}}$, it needs to compute a subset $\Ccal_{\alpha}(x)$. This is achieved by sorting the 
conformal scores and reusing computations across $\alpha$ values, which takes 
$\Ocal(m \log m + m n \log n)$ steps. 
Therefore, the overall time complexity is $\Ocal(m \log m + m n \log n)$.

\xhdr{Remarks} By using the MNL, we im\-plicitly assume the independence of irrelevant alternatives (IIA)~\citep{luce1959possible}, an axiom that states that the expert’{}s relative preference between two alternatives 
remains the same over all possible subsets containing these alternatives. 
While IIA is one of the most widely used axioms in the li\-te\-ra\-ture on discrete choice models, there is also a large body of experimental literature claiming to document real-world settings where IIA fails to hold~\citep{tversky1972elimination, huber1982adding, simonson1989choice}.
Fortunately, we have empirically found that experts may 
benefit from using our system even under strong violations 
of the IIA assumption in the estimator of the expert'{}s 
success pro\-ba\-bi\-li\-ty (\ie, when the estimator of
the expert'{}s success probability is not accurate), 
as shown in Figures~\ref{fig:robustness-synthetic} and~\ref{fig:robustness-real}.
% 
% Nevertheless, it would be very interesting to use more 
% sophisticated discrete choice models that do not rely on the IIA 
% assumption to accurately estimate the expert'{}s success 
% probability.

% \xhdr{Remarks} 
%
Conformal prediction is one of many possible ways to construct set-valued predictors~\citep{chzhen2021set}, \ie, predictors that, for each sample $x \in \Xcal$, output a set of label candidates $\Ccal(x)$.
In our work, we favor conformal predictors over alternatives because they 
provably output \emph{trustworthy} sets $\Ccal_{\alpha}(x)$ without making 
any assumption about the data distribution nor the classifier 
they rely upon.
In fact, we can use conformal predictors with any off-the-shelf classifier.
However, we would like to emphasize that our efficient search method (Algorithm~\ref{alg:near-optimal-alpha}) is rather generic and, together
with an estimator of the expert'{}s success probability with provable
guarantees,
% \footnote{\edit{Algorithm~\ref{alg:near-optimal-alpha} can use any estimator 
% $\hat{\mu}_{\alpha}$ such that, with probability at least $1-\delta$, it holds that 
% $|\hat{\mu}_{\alpha} - \PP[\hat Y = Y \,;\,  \Ccal_{\alpha}]| \leq \epsilon_{\alpha, 
% \delta}$ and is not affected by the type of the estimation error bound 
% $\epsilon_{\alpha, \delta}$.}}, 
may be used to find a near-optimal set-valued predictor within a discrete set of 
set-valued predictors that maximizes the expert'{}s success probability.
This is because our near-optimal guarantees in Proposition~\ref{prop:improve-error-practice} 
do not make use of the PAC guarantees afforded by conformal prediction, as discussed
previously. 
In Appendix~\ref{app:beyond-standard-cp}, we discuss an alternative set-valued 
predictor with PAC coverage guarantees, which may provide improved performance in scenarios
where the classifier underpinning our system has not particularly high average accuracy.
%
% \edit{
% In Appendix~\ref{app:topk}, we  we experiment with a 
% set-valued predictor that returns the 
% $k$ label values with the highest classifier scores, namely 
% a top-$k$ predictor, and show 
% the advantage of conformal predictors even over the optimal 
% top-k ones.
% }
%
We hope our work will encourage others to develop set-valued predictors 
specifically designed to serve decision support systems.

\begin{table}[t]
\caption{Empirical success probability % $\PP[\hat Y = Y \,;\, \Ccal_{\hat{\alpha}_1, \hat{\alpha}_2}]$ 
achieved by four different experts using our system during test, each with a 
different success pro\-ba\-bi\-li\-ty $\PP[\hat{Y}= Y \,;\, \Ycal]$, on four prediction tasks where the classifier achieves a different success probability $\PP[Y' = Y]$.
Each column corresponds to a prediction task, and each row to an expert.
In each task, the number of label values $n = 10$ and the size of the calibration and estimation sets is $m = 1{,}200$.
Each cell shows only the average since all standard errors are below $10^{-2}$.
} \label{table:synthetic-acc-10-labels}
\begin{center}
% \begin{small}
    % \small
    \begin{sc}
        \begin{tabular}{ccccr}
            \toprule
            \multirow{2}{*}{$\PP[\hat{Y}= Y \,;\, \Ycal]$} &\multicolumn{4}{c}{$\PP[Y' = Y]$}\\
            & 0.3 & 0.5 & 0.7 &0.9 \\
            \midrule
            0.3   &  0.41 &  0.58 &  0.75 &  0.91 \\
            0.5   &  0.55 &  0.68 &  0.80 &  0.93 \\
            0.7   &  0.72 &  0.79 &  0.87 &  0.95 \\
            0.9   &   0.90 &  0.91 &  0.95 &  0.98 \\
            % 0.3   &   0.4 &  0.57 &  0.74 &  0.91 \\
            % 0.5   &  0.55 &  0.68 &   0.80 &  0.93 \\
            % 0.7   &  0.72 &  0.79 &  0.88 &  0.95 \\ 
            % 0.9   &   0.90 &  0.92 &  0.95 &  0.98 \\
            \bottomrule
        \end{tabular}
    \end{sc}
\end{center}
% \vspace{-4mm}
\end{table}

% \vspace{-1mm}
% \section{Beyond Standard Conformal Prediction}
% \label{sec:modified-conformal-prediction}
% \vspace{-1mm}
% \input{050modified-cp}

\vspace{-1mm}
\section{Experiments on Synthetic Data}
\label{sec:synthetic}
% \vspace{-1mm}
In this section, we evaluate our system against the accuracy of the expert and the classifier, the size of the calibration and estimation sets, as well as the number of label va\-lues. 
Moreover, we analyze the robustness of our system to violations of the IIA assumption in the estimator of the expert'{}s success pro\-ba\-bi\-li\-ty\footnote{All algorithms ran on a Debian machine equipped with Intel Xeon E5-2667 v4 @ 3.2 GHz, 32GB memory and two M40 Nvidia Tesla GPU cards. See Appendix~\ref{app:implementation} for further details.}.

\begin{figure*}[t]
\centering
\subfloat[Empirical success probability vs $\alpha$]{\includegraphics[width=.5\columnwidth]{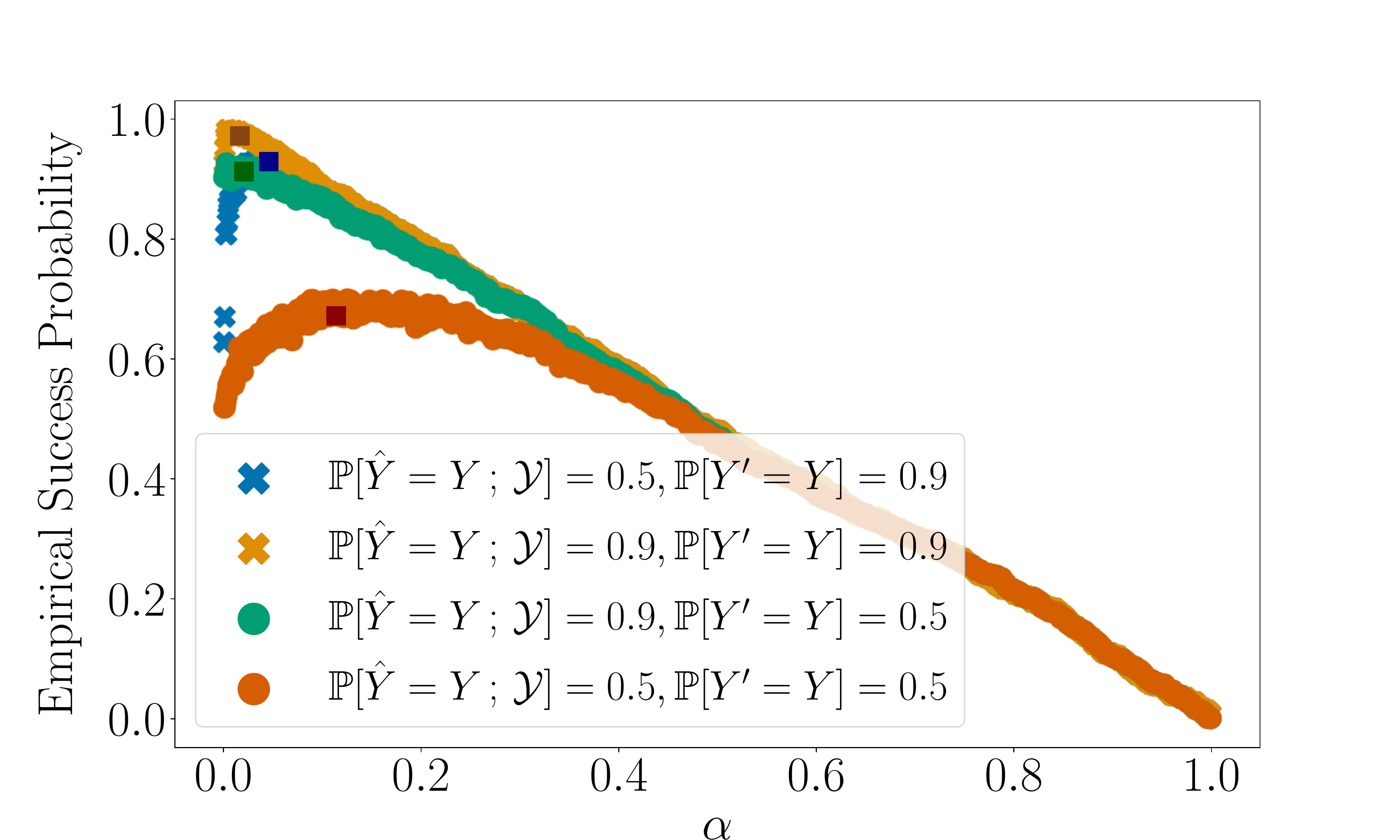}}
\subfloat[Empirical average set size vs $\alpha$]{\includegraphics[width=.5\columnwidth]{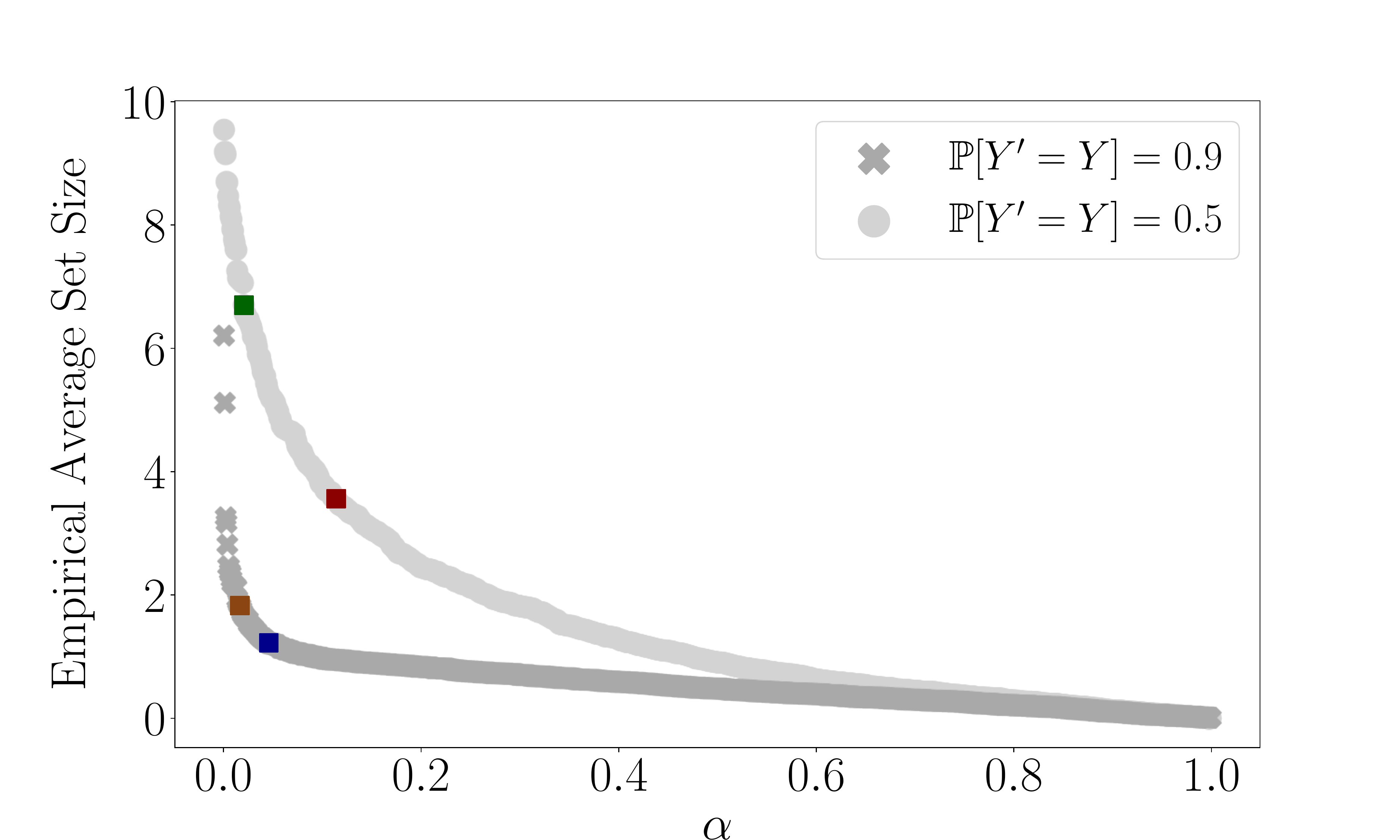}}
\caption{Empirical success probability achieved by two different experts using our system, each with a different success pro\-ba\-bi\-li\-ty $\PP[\hat{Y}= Y \,;\, \Ycal]$,
% $\PP[\hat{Y}= Y \,;\, \Ccal_{\alpha}]$ 
and average size of the recommended sets 
% $\EE{[|\Ccal_{\alpha}(X)|]}$ 
during test for each $\alpha \in \Acal$ on two synthetic prediction tasks where the classifier achieves a different success probability $\PP[Y' = Y]$.
Here, note that the empirical average set size only depends on the classifier'{}s success probability $\PP[Y' = Y]$, not the expert, and thus we only need two lines.  
In all experiments, the number of label values $n = 10$ and the size of the calibration and estimation sets is $m = 1{,}200$. 
Each marker corresponds to a different $\alpha$ value, and the darker points correspond to $\hat{\alpha}$. The coloring of the darker points for each prediction task is the same in both panels.} 
\label{fig:synthetic_error_alpha}
%
% \vspace{-3mm}
\end{figure*}

\xhdr{Experimental setup} 
We create a variety of synthetic prediction tasks, each with $20$ features per sample and a varying\- number of label values $n$ and difficulty. 
Refer to Appendix~\ref{app:implementation} for more details about the prediction tasks.
For each prediction task, we generate $10{,}000$ samples, pick $20$\% of these samples at random as test set, which we use to estimate the performance of our system, 
and also randomly split the remaining $80$\% into three disjoint subsets for 
training, calibration, and estimation, whose sizes we vary across experiments. 
In each experiment, we specify the number of samples in the calibration and 
estimation sets---the remaining samples are used for training.

For each prediction task, we train a logistic re\-gression model $P_{\theta}(Y' \given X)$, which depending on the difficulty of the prediction task, achieves different success probability values $\PP[ Y' = Y ]$. 
Moreover, we sample the expert'{}s predictions $\hat Y$ from the multinomial logit model defined by Eq.~\ref{eq:mnl-pred-prob}, with $C_{yy} = \frac{\pi}{n} \pm \gamma\epsilon_c$ and $C_{yy'} = \frac{1 - C_{yy}}{n} \pm \beta$, 
where $\pi$ is a parameter that controls the expert'{}s success probability $\PP[ \hat{Y} = Y \,;\, \Ycal]$, 
$\epsilon_c \sim \text{U}(0, \min(1 - \frac{\pi}{n},  \frac{\pi}{n}))$, 
$\beta\sim \text{N}(0, ((1 - C_{yy})/(6n))^2)$ for all $y \neq y'$,
and $\gamma$ is a normalization term. 
Finally, we repeat each experiment ten times and, each time, we sample different train, estimation, calibration, and test sets following the above procedure.

\xhdr{Experts always benefit from our system even if the classi\-fier has low accuracy}
We estimate the success probability $\PP[\hat{Y} = Y \,;\, \Ccal_{\hat{\alpha}}]$ achieved by four diffe\-rent experts, each with a different success probability $\PP[\hat{Y}= Y \,;\, \Ycal]$, on four prediction tasks where the classifier achieves a different success probability $\PP[Y' = Y]$. 
Table~\ref{table:synthetic-acc-10-labels} summarizes the results, where each column corresponds to a different prediction task and each row corresponds to a different expert.
We find that, using our system, the expert solves the prediction task significantly more accurately than the expert or the classifier on their own.
Moreover, it is rather remarkable that, even if the classifier has low accuracy, the expert always benefits from using 
our system---in other words, our system is robust to the performance of the classifier it relies on.
In Appendix~\ref{app:results}, we show qualitatively similar results for prediction tasks with other values of $n$ and $m$.
%
% manuel: we now refer to the appendix below in the section of performance and average set size vs. \alpha
%
% , and, in 
% Appendix~\ref{app:set-size-distr}, we show that, the smaller the near optimal $\hat \alpha$, the greater 
% the spread of the empirical distribution of the size of the subsets $\Ccal_{\hat \alpha}(X)$.
%
% Finally, since we found that, in the majority of repetitions 
% of each experiment, % , for each prediction task, 
% the near-optimal $\hat{\alpha}_2 = 1$,
% % \scriptsize In the prediction tasks with $\PP[Y'=Y] = 0.7$,
% % $\hat{\alpha}_2 \neq 1$ occured a bit more frequently, 
% % though these cases were still a minority.}}
% % 
% in what follows, we only experiment with systems 
% $\Ccal_{\hat{\alpha}} = \Ccal_{\hat{\alpha}, 1}$ that construct $\Ccal_{\hat{\alpha}}(X)$ using Eq.~\ref{eq:prediction-set}. 

% 
% manuel: I think it is more important to include the results about violations of IIA than the results
% below. I think it is best to have an appendix where we show the figures and text below also together
% for synthetic and real data in the Appendix and we refer to it from the section where we introduce the
% optimization method.
%
\xhdr{The performance of our system under $\hat{\alpha}$ found by Algorithm~\ref{alg:near-optimal-alpha} and 
under $\alpha^{*}$ is very similar}
Given three prediction tasks where the expert and the classifier achieve diffe\-rent success probabilities $\PP[\hat{Y} = Y \,;\, \Ycal]$ and $\PP[ Y' = Y ]$,
we compare the performance of our system under the near-optimal $\hat{\alpha}$ found by Algorithm~\ref{alg:near-optimal-alpha} and under all other possible $\alpha \in \Acal$ values. %, including the optimal $\alpha^{*}$. 
Figure~\ref{fig:synthetic_error_alpha} summarizes the results, which suggest that: 
(i) the performance under $\hat{\alpha}$ is very close to that under $\alpha^{*}$, as suggested by Proposition~\ref{prop:improve-error-practice}; and,
(ii) as long as $\alpha \leq \alpha^*$, the performance of our system increases monotonically with respect to $\alpha$, however, once $\alpha > \alpha^{*}$, the performance deteriorates as we increase $\alpha$.
(iii) the higher the expert'{}s success probability $\PP[ \hat{Y} = Y \,;\, \Ycal]$, the smaller the near optimal $\hat \alpha$ and thus the greater the average size of the subsets $\Ccal_{\hat \alpha}(X)$. 
In Appendix~\ref{app:set-size-distr}, we also show that, the smaller the near optimal $\hat \alpha$, the greater the spread 
of the empirical distribution of the size of the subsets $\Ccal_{\hat \alpha}(X)$.
We found qualitatively similar results using\- other expert-classifier pairs with different success pro\-ba\-bi\-li\-ties.

\xhdr{Our system needs a relatively small amount of calibration and estimation data}
We vary the amount of calibration and estimation data $m$ we feed into Algorithm~\ref{alg:near-optimal-alpha} and, each time, estimate the expert'{}s success probability $\PP[\hat Y = Y \,;\, \Ccal_{\hat{\alpha}}]$.
Across prediction tasks, we consistently find that our system needs a relatively small amount of ca\-li\-bra\-tion and estimation data to perform well.
For exam\-ple, for all prediction tasks with $n = 10$ label values and va\-rying level of difficulty, 
% if we increase the amount of calibration and estimation data from $m = 160$ to $m = 1{,}200$, 
the relative gain in empirical success probability achieved by
an expert using our system with respect to an expert on their own, averaged across experts with $\PP[\hat Y = Y \,;\, \Ycal] \in \{0.3, 0.5, 0.7, 0.9\}$, goes from $47.56 \pm 4.51 \%$ for $m = 160$ to $48.66 \pm 4.54\%$ for $m = 1{,}200$. %
%

% \vspace{-2mm}
\xhdr{The greater the number of label values, the more an expert benefits from using our system}
We consider prediction tasks with a varying number of label values, from $n = 10$ to $n = 100$, and estimate the expert'{}s success pro\-ba\-bi\-lity $\PP[\hat Y = Y \,;\, \Ccal_{\hat{\alpha}}]$ for each task.
Our results suggest that the relative gain in success probability, 
% achieved by an expert using our system with respect to an expert on their own,
% $\PP[\hat{Y}= Y \,;\, \Ccal_{\hat{\alpha}}]$ with respect to 
% $\PP[\hat{Y}= Y \,;\, \Ycal]$, 
averaged across experts with $\PP[\hat Y = Y \,;\, \Ycal] \in \{0.3, 0.5, 0.7, 0.9\}$, increases with the number of label values.
For example, for $m = 400$, it goes from $48.36 \pm 4.50\%$ for $n = 10$ to $69.44\pm 5.20\%$ for $n = 100$. For other $m$ values, 
we found a similar trend. 
\begin{figure*}[t]
\centering
\captionsetup[subfloat]{labelformat=empty,textfont=small}
\foreach \human in {0.5, 0.7, 0.9}{
        % \foreach \machine in {0.5,0.7,0.9}{
\subfloat[\qquad\qquad\quad$\PP{[\hat{Y}= Y\,;\, \Ycal]} = \human$]{
            \includegraphics[width=0.32\linewidth]{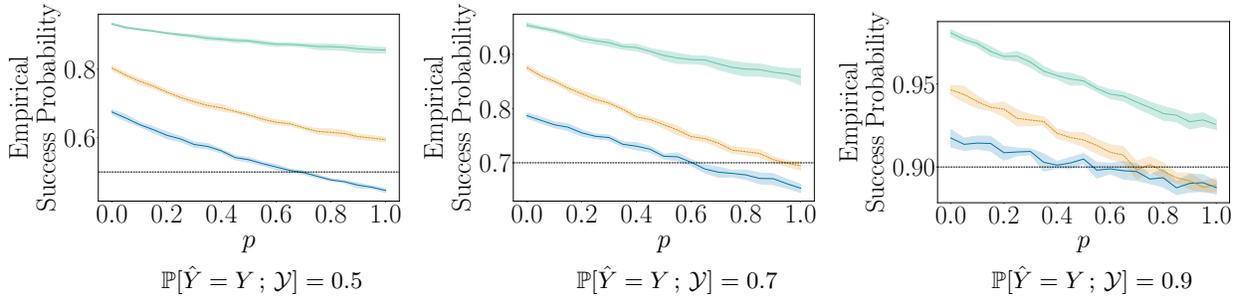}}}
    % \centering            \includegraphics[width=0.33\linewidth]{plots/robustness_synthetic_human\human.pdf}}}

    % \subfloat[\qquad\qquad\qquad$\PP{[\hat{Y}= Y\,;\, \Ycal]} = 0.9$]{ \includegraphics[width=0.5\linewidth]{plots/robustness_synthetic_human0.9.pdf}}

\caption{Empirical success probability achieved by three different experts using our system during test, each with a different success probability $\PP[\hat Y = Y \,;\, \Ycal]$, against severity $p$ of the violation of the IIA assumption on three prediction tasks where the classifier achieves a different success probability $\PP[Y' = Y]$.
In each panel, the horizontal line shows the empirical success probability achieved by the expert at solving the (original) multiclass task during test. 
The number of labels is $n=10$ and the size of the calibration and estimation sets is $m=1{,}200$. Shaded regions correspond to $95\%$ confidence intervals.}
\label{fig:robustness-synthetic}
% \vspace{-3mm}
\end{figure*}

\xhdr{Our system is robust to strong violations of the IIA assumption in the estimator of the
expert'{}s success probability}
To study the robustness of our system to violations of the IIA assumption in the estimator of the expert'{}s success pro\-ba\-bi\-li\-ty, we allow the expert'{}s preference $u_{y y'}$ for each label 
value $y' \neq y$ in Eq.~\ref{eq:mnl-pred-prob} to depend on the corresponding prediction set $\Ccal_{\hat{\alpha}}(x)$ at test time.
More specifically, we set 
\begin{equation*}
u_{y y'} = \log \left (C_{yy'} + p \, \frac{\II[y' \neq y]}{|C_{\hat{\alpha}}(x)\backslash\{y\}|} \sum_{y'' \notin C_{\hat{\alpha}}(x)} C_{yy''} \right),
\end{equation*}
where $p \in [0, 1]$ is a parameter that controls the severity of the violation
of the IIA assumption at test time.
Here, note that if $p = 1$, the expert does not benefit from using our system as long as the prediction set $C_{\alpha}(x) \neq \{y\}$, \ie, the expert'{}s conditional success probability is given by $\PP[\hat Y = y \,;\, \Ccal_{\alpha} \given y \in \Ccal_{\alpha}(x)] = \PP[\hat Y = y \,;\, \Ycal]$.
Figure~\ref{fig:robustness-synthetic} summarizes the results, which show that our system is robust to (strong) violations of the IIA assumption in the estimator of the expert'{}s success probability.

\begin{table}[t]
\caption{Empirical success probabilities 
% $\PP[Y' = Y]$ and $\PP[\hat Y = Y \,;\, \Ccal_{\hat{\alpha}}]$ 
achieved by three popular deep neural network classifiers and by an  expert 
using our system ($\Ccal_{\hat{\alpha}}$) and the best top-$k$ set-valued predictor ($\Ccal_k$) with these classifiers during test on the CIFAR-10H dataset.
The size of the calibration and estimation sets is $m = 1{,}500$ and the expert's empirical success probability at solving the (original) multiclass task is $\PP[\hat{Y}= Y \,;\, \Ycal] \approx 0.947$.
Each cell shows only the average since the standard errors are all 
below $10^{-2}$.} \label{table:real-acc-labels}
\begin{center}
% \begin{small}
    % \small
    \begin{sc}
        \begin{tabular}{lccc}
            \toprule
            \small
            & Classifier & $\Ccal_{\hat{\alpha}}$ & $\Ccal_k$ \\
            \midrule
            ResNet-110     & 0.928 &  0.987 & 0.967 \\
            PreResNet-110    & 0.944 & 0.989 & 0.972 \\
            DenseNet      & 0.964 & 0.990 & 0.980  \\
            \bottomrule
        \end{tabular}
    \end{sc}
% \end{small}
\end{center}
% \vspace{-4mm}
\end{table}

\vspace{-1mm}
\section{Experiments on Real Data}
\label{sec:real}
% \vspace{-1mm}
In this section, we experiment with a dataset with real expert predictions on a 
multiclass classification task over natural ima\-ges and several popular and highly 
accurate deep neural network classifiers.
In doing so, we benchmark the performance of our system against a competitive top-$k$ 
set-valued predictor baseline, which always returns the $k$ label values with the 
highest scores, and analyze its robustness to violations of the IIA assumption in 
the estimator of the expert'{}s success pro\-ba\-bi\-li\-ty.
% \footnote{Here, we focus on systems $\Ccal_{\alpha} = \Ccal_{\alpha, 1}$ % that construct $\Ccal_{\alpha}(X)$ using Eq.~\ref{eq:prediction-set} 
% because, whenever the classifiers are highly accurate, systems $\Ccal_{\alpha_1, \alpha_2}$ with $\alpha_2 \neq 1$ do not offer a competitive advantage.}.
%
Here, we would like to explicitly note that we rely on the confusion matrix estimated
using real expert predictions on the (original) multiclass classification task and
the multinomial logit model defined by Eq.~\ref{eq:mnl-pred-prob} to estimate the performance of our system and the competitive top-$k$ set-valued predictor 
baseline---no real experts actually used our decision support system.

\xhdr{Data description}
We experiment with the dataset CIFAR-10H~\citep{peterson2019human}, which contains $10{,}000$ natural images taken from the test set of the standard CIFAR-10~\citep{krizhevsky2009learning}.
Each of these images belongs to one of $n = 10$ classes and contains approximately $50$ expert predictions $\hat Y$\footnote{The dataset CIFAR-10H is among the only publicly available datasets (released under Creative Commons BY-NC-SA 4.0 license) that we found containing multiple expert predictions per sample, necessary to estimate $\bf C$, a relatively large number of samples, and more than two classes.
However, since our methodology is rather general, our system may be useful in other applications.}.
Here, we randomly split the dataset into three disjoint subsets for calibration, estimation and test, whose sizes we vary across experiments.
In each experiment, we use the test set to estimate the performance of our system 
and we specify the number of samples in the calibration and estimation sets---the re\-mai\-ning samples are used for testing.

\xhdr{Experimental setup} Rather than training a classifier, we use three popular and highly accurate deep neural network classifiers trained on CIFAR-10, namely ResNet-110~\citep{he2016deep}, PreResNet-110~\citep{he2016identity} and DenseNet~\citep{huang2018densely}.
Moreover, we use the human predictions $\hat{Y}$ to estimate the confusion matrix $\bf C$ for the 
%
% manuel: the confusion matrix refers to the original classificaiton task where we
% truly have real expert predictions.
%
% \edit{simulated} 
%
expert predictions in the (original) multiclass classification task~\citep{kerrigan2021combining} and then sample the % \edit{simulated} 
expert'{}s prediction $\hat Y$ from the multinomial logit model defined by Eq.~\ref{eq:mnl-pred-prob} to both estimate the % \edit{simulated} 
expert'{}s conditional success probabilities in Eq.~\ref{eq:hat-mu} in Algorithm~\ref{alg:near-optimal-alpha} and estimate the % \edit{simulated} 
expert'{}s success probability during testing. 
%
% \edit{
In what follows, even though the expert'{}s performance during testing is estimated 
using the multinomial logit model, rather than using real predictions from experts 
using our system, 
we refer to (the performance of) such a \emph{simulated} expert as an expert.
% }
% 

\xhdr{Performance evaluation}
We start by estimating the success pro\-ba\-bi\-lity $\PP[\hat Y = Y \,;\, \Ccal_{\hat{\alpha}}]$ achieved by an expert using our system ($\Ccal_{\hat \alpha}$) and the best top-$k$ set-valued predictor ($\Ccal_{k}$), which returns the $k$ label values with the highest scores\footnote{ Appendix~\ref{app:topk} shows the success probability achieved by an expert using the top-$k$ set-valued predictor for different $k$ values both for synthetic and real data.}.
Table~\ref{table:real-acc-labels} summa\-rizes the results, where we also report the (empirical) success probability achieved by an expert solving the (original) multiclass task in their own.
We find that, by allowing for recommended subsets of varying size, our system is consistently superior to the top-$k$ set-valued predictor.
Moreover, we also find it very encouraging that, although the classifiers are highly accurate, our results suggest that an expert using our system can solve the prediction task significantly more accurately than the classifiers.
More spe\-ci\-fi\-cally, the relative reduction in misclassification probability goes from $72.2$\% (DenseNet) to $81.9$\% (ResNet-110).
Finally, by using our system, our results suggest that the (average) expert would reduce their misclassification probability by $\sim$$80$\%.

\begin{figure}[t]
    \centering
    \captionsetup[subfloat]{labelformat=empty}
    \includegraphics[scale=.2]{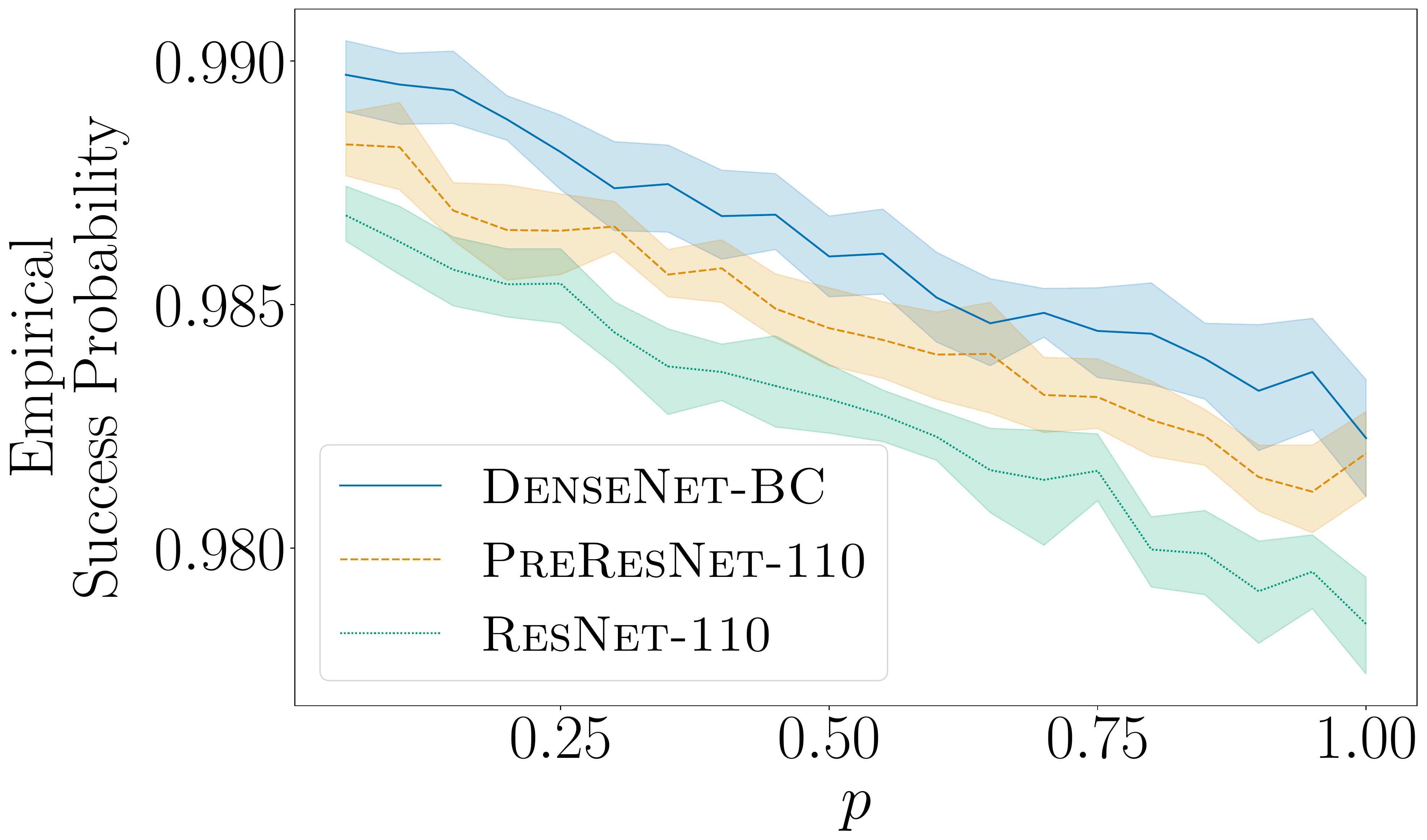}
    % \vspace{-2mm}
    \caption{Empirical success probability achieved by an expert using our system with three different classifiers during test against severity $p$ of the violation of the IIA assumption on the CIFAR-10H dataset.
    The empirical success probability achieved by the expert at solving the (original) multiclass task during test is $\PP[\hat Y = Y \,;\, \Ycal] \approx 0.947$. The size of the calibration and estimation sets is $m=1{,}500$. Shaded regions correspond to $95\%$ confidence intervals.}
    \label{fig:robustness-real}
    % \vspace{-4mm}
\end{figure}

\xhdr{Robustness to violations of the IIA assumption
in the estimator of the expert'{}s success probability} 
To study the robustness of our system to violations of the IIA assumption in the estimator of the expert'{}s success pro\-ba\-bi\-li\-ty, we use the same experimental setting as in the synthetic experiments, where the parameter $p$ controls the severity of the violation of the IIA assumption at test time.
%
% To analyze the sensitivity of our results to the accuracy of the estimator of the 
% expert'{}s success probability, 
%
%
Figure~\ref{fig:robustness-real} summarizes the results for different $p$ values. It is remarkable that, even for highly accurate classifiers like the ones used for our experiments, the expert benefits from using our system even when $p=1$. This is because, for accurate classifiers, many prediction sets are singletons containing the true label, as shown in Appendix~\ref{app:alpha-error}. 

\vspace{-2mm}
\section{Conclusions}
\label{sec:conclusion}
% \vspace{-2mm}
We have initiated the development of automated decision
support systems that, by design, do not require human experts to 
understand when each of their recommendations is accurate to improve their 
performance with high pro\-ba\-bi\-lity.
We have focused on multiclass classi\-fi\-ca\-tion and 
designed a system that, for each data sample, recommends a subset of 
labels to the experts using a classi\-fier.
Moreover, we have shown that our system can help experts make predictions 
more accurately and is robust to the accuracy of the classi\-fier and the
estimator of the expert'{}s success probability.

Our work opens up many interesting avenues for future work. 
For example, we have considered a simple, well-known conformal score function 
from the literature.
However, it would be valuable to develop score functions especially designed 
for decision support systems.
Moreover, it would be interesting to perform online estimation of the expert'{}s 
conditional success probability.
Further, it would be important to investigate the ethical impact of our system, 
including human trust and bias, understand the robustness of our system to malicious
attacks, and consider alternative performance metrics such as expert prediction time. 
Finally, it would be important to deploy and evaluate our system on a real-world 
application with human experts.
%
%

% \vspace{2mm}
\section*{Acknowledgements} We would like to thank the anonymous reviewers for constructive feedback, which has helped improve our paper. 
Gomez-Rodriguez acknowledges support from the European Research Council (ERC) under the European Union'{}s Horizon 2020 research and innovation programme (grant agreement No. 945719). Wang acknowledges support from NSF Awards IIS-1901168 and IIS-2008139. All content represents the opinion of the authors, which is not necessarily shared or endorsed by their respective employers and/or sponsors. 
{ 
\bibliographystyle{apalike}
\bibliography{prediction-sets}
}

\clearpage
\newpage

\appendix

\section{Proofs}

\subsection{Proof of Proposition~\ref{prop:improve-error-practice}}
Given the estimators $\hat{\mu}_{\alpha}$ of $\PP[\hat{Y} = Y \,;\, \Ccal_\alpha]$, we have that, for each $\alpha \in \Acal$, it holds that
\begin{equation}
\label{eq:event_bounded_error}
\left|\hat{\mu}_\alpha - \PP[\hat{Y} = Y \,;\, \Ccal_\alpha]\right|\leq \epsilon_{\alpha, \delta/m}
\end{equation}
with probability at least $1 - \delta/m$. By applying the union bound, we know that the above events hold simultaneously for all $\alpha \in \Acal$ with probability at least $1 - \delta$.  Moreover, by rearranging, the above expression can be
rewritten as
\begin{align}\label{eq:estimation-error-bound}
\hat{\mu}_{\alpha} - \epsilon_{\alpha, \delta /m} \leq \PP[\hat{Y} = Y \,;\, \Ccal_{\alpha}]
 \leq \hat{\mu}_{\alpha} + \epsilon_{\alpha, \delta / m}. 
\end{align}
Let $\hat{\alpha} = \argmax_{\alpha \in \Acal}\{(\hat{\mu}_{\alpha} - \epsilon_{\alpha, \delta / m})\}$. For $\hat{\alpha}$, with probability $1 - \delta$, it holds that for all $\alpha \in \Acal$,
\begin{align*}
\PP[\hat{Y} = Y \,;\, \Ccal_{\hat{\alpha}}]
\geq \hat{\mu}_{\hat{\alpha}} - \epsilon_{\hat{\alpha}, \delta/m}
\geq \hat{\mu}_{\alpha} - \epsilon_{\alpha, \delta / m} &= \hat{\mu}_{\alpha} - \epsilon_{\alpha, \delta / m} +2\epsilon_{\alpha, \delta / m} - 2\epsilon_{\alpha, \delta / m}\\
& = \hat{\mu}_{\alpha} + \epsilon_{\alpha, \delta / m} - 2\epsilon_{\alpha, \delta / m}, \\
& \geq \PP\sbr{\hat{Y} = Y; \Ccal_{\alpha}} - 2 \epsilon_{\alpha, \delta / m},
\end{align*}
where the last inequality follows from Eq.~\ref{eq:estimation-error-bound}. 

\subsection{Derivation of Error Expression for Hoeffding'{}s Inequality} \label{app:hoeffding}
From Hoeffding's inequality we have that:
\begin{theorem}
Let $Z_1,..., Z_k$ be i.i.d., with $Z_i \in [a, b], i = 1,...,k, a < b$ and $\hat{\mu}$ be the empirical estimate $\hat{\mu} = \frac{\sum_{i = 1}^{k}Z_i}{k}$ of $\EE[Z] = \EE[Z_i]$. Then:
\begin{align}
    \PP \left[ \hat{\mu} - \EE[Z] \geq \epsilon \right] \leq \exp{\left(- \frac{2k\epsilon^2}{(b - a)^2} \right)}
\end{align}
and
\begin{align}
    \PP \left[ \hat{\mu} - \EE[Z] \leq -\epsilon \right] \leq \exp{\left(- \frac{2k\epsilon^2}{(b - a)^2} \right)}
\end{align}
hold for all $\epsilon \geq 0$.
\end{theorem}

In our case we have $k = m$ and % $Z_i = \PP\sbr{\hat Y = Y_i \given \Ccal_{\alpha}(X_i), Y_i} \in (0, 1)$ \lwedit{
$Z_i = \II\cbr{Y_i \in \Ccal_{\alpha}(X_i)}\PP\sbr{\hat{Y} = Y_i;\Ccal_{\alpha} \given Y_i \in \Ccal_{\alpha}(X_i)} \in (0, 1)$. 
Moreover, note that the expectation of $Z_i$ is given by:
\begin{align*}
\EE[Z_i]
&= \EE\sbr{\II\cbr{Y_i \in \Ccal_{\alpha}(X_i)}\PP\sbr{\hat{Y} = Y_i;\Ccal_{\alpha} \given Y_i \in \Ccal_{\alpha}(X_i)}}\\
&=\EE\sbr{\II\cbr{Y_i \in \Ccal_{\alpha}(X_i)}\PP\sbr{\hat{Y} = Y_i;\Ccal_{\alpha} \given Y_i \in \Ccal_{\alpha}(X_i)} + \II\cbr{Y_i \notin \Ccal_{\alpha}(X_i)} \PP\sbr{\hat{Y} = Y_i; \Ccal_{\alpha} \given Y_i \notin \Ccal_{\alpha}(X_i)}}\\
&=\EE\sbr{\PP\sbr{\hat{Y} = Y_i; \Ccal_{\alpha}}}\\
&=\PP\sbr{\hat{Y} = Y_i ; \Ccal_{\alpha}}, 
\end{align*}
where the expectations are over the joint distribution of prediction sets $\Ccal_{\alpha}(X)$ and true labels $Y$.

% , with $y_i \in \Ccal_{\alpha}(x_i)$, for $i = 1, ..., k_{\alpha}$. 
Hence, for the empirical estimate $\hat{\mu} = \hat{\mu}_{\alpha}$ of $\PP[\hat Y = Y \,;\,  \Ccal_{\alpha}]$ and its error $\epsilon=\epsilon_{\alpha, \delta}$:
\begin{align}
    \PP \left[ \hat{\mu}_{\alpha} - \PP[\hat Y = Y \,;\,  \Ccal_{\alpha}] \geq \epsilon_{\alpha, \delta} \right] \leq \exp{\left(- \frac{2m\epsilon_{\alpha, \delta}^2}{(1 - 0)^2} \right)}
\end{align}
and 
\begin{align}
    \PP \left[ \hat{\mu}_{\alpha} - \PP[\hat Y = Y \,;\,  \Ccal_{\alpha}] \leq -\epsilon_{\alpha, \delta} \right] \leq \exp{\left(- \frac{2m\epsilon_{\alpha, \delta}^2}{(1 - 0)^2} \right)}
\end{align}
hold. Further, if we set 
\begin{align}
\label{eq:delta}
\delta = \exp{\left(- 2m\epsilon_{\alpha, \delta}^2 \right)},
\end{align}
then 
\begin{align}
\label{eq:lower-bound-expectation}
1 - \PP \left[ \hat{\mu}_{\alpha} - \PP[\hat Y = Y \,;\,  \Ccal_{\alpha}] \leq \epsilon_{\alpha, \delta} \right] \leq \delta \Rightarrow \PP \left[ \hat{\mu}_{\alpha} - \PP[\hat Y = Y \,;\,  \Ccal_{\alpha}] \leq \epsilon_{\alpha, \delta} \right] \geq 1 - \delta
\end{align}
and 
\begin{align}
\label{eq:upper-bound-expectation}
1 - \PP \left[ \hat{\mu}_{\alpha} - \PP[\hat Y = Y \,;\,  \Ccal_{\alpha}] \geq -\epsilon_{\alpha, \delta} \right] \leq \delta \Rightarrow \PP \left[ \hat{\mu}_{\alpha} - \PP[\hat Y = Y \,;\,  \Ccal_{\alpha}] \geq -\epsilon_{\alpha, \delta} \right] \geq 1 - \delta
\end{align}
hold for any $\epsilon_{\alpha, \delta} \geq 0$. As follows, based on Eq.~\ref{eq:delta}:
\begin{align*}
    \delta = \exp{\left(- 2m\epsilon_{\alpha, \delta}^2 \right)}\Rightarrow
    \log{\frac{1}{\delta}} =  2m\epsilon_{\alpha, \delta}^2 \Rightarrow 
   \epsilon_{\alpha, \delta}^2  = \frac{\log{\frac{1}{\delta}}}{2m} &\Rightarrow
   \epsilon_{\alpha, \delta} = \sqrt{\frac{\log{\frac{1}{\delta}}}{2m}}.
\end{align*}

\subsection{Proof of Proposition~\ref{prop:modified-cp-beta}}
We proceed similarly as in the Appendix A.5 in~\citet{hulsman2022distribution}. 
First, note that, by definition, we have that
\begin{equation*}
\hat{q}_{\alpha_1} = s_{(\lceil(1-\alpha_1)(m+1)\rceil)} \quad \text{and} \quad \hat{q}_{\alpha_2} = s_{(\lceil(1-\alpha_2)(m+1)\rceil)},
\end{equation*}
where $s_{(i)}$ denotes the $i$-th smallest conformal score in the calibration set $\Dcal_{\text{cal}}$.
Then, as long as the conformal scores in the calibration set are almost surely distinct, it follows directly from Proposition 4 in~\citet{hulsman2022distribution} that
\begin{equation}\label{eq:modified-cp-order-stat-beta}
    \PP\left[\hat{q}_{\alpha_2} < s(X,Y) \leq  \hat{q}_{\alpha_1} \given \Dcal_{\text{cal}} \right] \sim
    \text{Beta}(l, m - l + 1),
\end{equation}
where $l=\lceil(m+1)(1 - \alpha_1)\rceil - \lceil(m+1)(1 - \alpha_2)\rceil$.
Moreover, for any $(X, Y) \sim P(X)P(Y \given X)$, we have that, by construction, $Y \in \Ccal_{\alpha_1, \alpha_2}(X)$ if and only if 
$s(X,Y) \in (\hat{q}_{\alpha_2}, \hat{q}_{\alpha_1})$. % (s_{(\lceil(1-\alpha_2)(m+1)\rceil)}, s_{(\lceil(1-\alpha_1)(m+1)\rceil)}]$.
Then, Eq.~\ref{eq:modified-cp-beta} follows directly from Eq.~\ref{eq:modified-cp-order-stat-beta}.
% }
\clearpage
\newpage

\section{Implementation Details} \label{app:implementation}
To implement our algorithms and run all the experiments on synthetic and real data, we used PyTorch 1.12.1, NumPy 1.20.1 and Scikit-learn 1.0.2 on Python 3.9.2. 
For reproducibility, we use a fixed random seed in all random procedures. Moreover, we set $\delta = 0.1$ everywhere.

\xhdr{Synthetic prediction tasks} 
We create $4 \times 3 = 12$ different prediction tasks, where
we vary the number of labels $n \in \{10,50,100\}$ and the
level of difficulty of the task.
More specifically, for each value of $n$, we create four different tasks of increasing difficulty where the success probability of the logistic regression classifier is $\PP[Y'=Y] = 0.9$, $0.7$, $0.5$ and $0.3$, respectively.

To create each task, we use the function \texttt{make\_classification} of the Scikit-learn library. 
This function allows the creation of data for synthetic prediction tasks with very particular user-defined characteristics, through the generation of clusters of normally distributed points on the vertices of a multidimensional hypercube. 
The number of the dimensions of the hypercube indicates the number of informative features of each sample, which in our case we set at 15 for all prediction tasks. 
Linear combinations of points, \ie, the informative features, are used to create redundant features, the number of which we set at 5. 
The difficulty of the prediction task is controlled through the size of the hypercube, with a multiplicative factor, namely \texttt{clas\_sep}, which we tuned accordingly for each value $n$ so that the success probability of the logistic regression classifier above spans a wide range of values across tasks. 
All the selected values of this parameter can be found in the configuration file \texttt{config.py} in the code. 
Finally, we set the proportion of the samples assigned to each label, \ie, the function parameter \texttt{weights}, using a Dirichlet distribution of order $n$ with parameters $\alpha_1=...=\alpha_n=1$.

\clearpage
\newpage 

\section{Additional Synthetic Prediction Tasks, Number of  Labels and Amount of Calibration and Estimation Data}
\label{app:results}
To complement the results in Table~\ref{table:synthetic-acc-10-labels} in the main paper, we experiment with additional prediction tasks with different number of labels $n$ and amount of calibration and estimation data $m$.
For each value of $n$ and $m$, we estimate the success probability $\PP[\hat Y = Y \,;\, \Ccal_{\hat{\alpha}}]$ achieved by four different experts using our system, each with a different success probability $\PP[\hat{Y}= Y\,;\, \Ycal]$, on four prediction tasks where the classifier achieves a different success probability $\PP[Y' = Y]$.
Figure~\ref{table:synthetic-acc-additional} summarizes the results.
%
% m=1200
% n = 50
\begin{figure}[h]
\centering
\subfloat[$n=50$, $m=1{,}200$ ]{
% \begin{table}
% \begin{center}
% \begin{small}
% \begin{sc}
\begin{tabular}{ccccr}
\toprule
\multirow{2}{*}{$\PP[\hat{Y}= Y\,;\, \Ycal]$} 
&\multicolumn{4}{c}{$\PP[Y' = Y]$}\\
& 0.3 & 0.5 & 0.7 &0.9 \\
\midrule
0.3   &  0.56 &  0.72 &  0.84 &  0.94 \\
0.5   &  0.68 &   0.80 &  0.89 &  0.95 \\
0.7   &  0.79 &  0.87 &  0.93 &  0.97 \\
0.9   &  0.92 &  0.95 &  0.97 &  0.99 \\
% 0.3   &  0.56 &  0.72 &  0.85 &  0.93 \\
% 0.5   &  0.67 &   0.80 &  0.89 &  0.96 \\
% 0.7   &  0.79 &  0.87 &  0.93 &  0.97 \\
% 0.9   &  0.91 &  0.95 &  0.97 &  0.99 \\
\bottomrule
\end{tabular}
% \end{sc}
% \end{small}
% \end{center}
% \end{table}
}
\hspace{4pt}
\subfloat[$n=100$, $m=1{,}200$ ]{
% \begin{table}
% \begin{center}
\begin{tabular}{ccccr}
\toprule
\multirow{2}{*}{$\PP[\hat{Y}= Y\,;\, \Ycal]$} 
&\multicolumn{4}{c}{$\PP[Y' = Y]$}\\
& 0.3 & 0.5 & 0.7 &0.9 \\
\midrule
0.3   &  0.62 &  0.76 &  0.87 &  0.95 \\
0.5   &  0.72 &  0.83 &  0.91 &  0.96 \\
0.7   &  0.83 &   0.90 &  0.95 &  0.98 \\
0.9   &  0.93 &  0.96 &  0.98 &  0.99 \\
% 0.3   &  0.63 &  0.76 &  0.88 &  0.95 \\
% 0.5   &  0.73 &  0.84 &  0.91 &  0.97 \\
% 0.7   &  0.82 &   0.90 &  0.95 &  0.98 \\
% 0.9   &  0.93 &  0.95 &  0.98 &  0.99 \\
\bottomrule
\end{tabular}
% \end{center}
% \end{table}
}
\\
\vspace{7pt}
% m=400, n=10
\subfloat[$n=10$, $m=400$]{
\begin{tabular}{ccccr}
\toprule
\multirow{2}{*}{$\PP[\hat{Y}= Y\,;\, \Ycal]$} 
&\multicolumn{4}{c}{$\PP[Y' = Y]$}\\
& 0.3 & 0.5 & 0.7 &0.9 \\
\midrule
0.3   &   0.42 &  0.58 &  0.75 &  0.91 \\
0.5   &  0.55 &  0.66 &   0.80 &  0.93 \\
0.7   &  0.72 &  0.79 &  0.87 &  0.96 \\
0.9   &  0.90 &  0.92 &  0.94 &  0.98 \\
% 0.3   &   0.40 &  0.57 &  0.75 &  0.88 \\
% 0.5   &  0.55 &  0.68 &   0.80 &   0.90 \\
% 0.7   &  0.72 &  0.79 &  0.88 &  0.93 \\
% 0.9   &   0.90 &  0.92 &  0.95 &  0.97 \\
\bottomrule
\end{tabular}}
\hspace{4pt}
% m=400, n=50
\subfloat[$n=50$, $m=400$]{\begin{tabular}{ccccr}
\toprule
\multirow{2}{*}{$\PP[\hat{Y}= Y\,;\, \Ycal]$} 
&\multicolumn{4}{c}{$\PP[Y' = Y]$}\\
& 0.3 & 0.5 & 0.7 &0.9 \\
\midrule
0.3   &  0.56 &  0.73 &  0.84 &  0.94 \\
0.5   &  0.67 &   0.80 &  0.88 &  0.96 \\
0.7   &   0.79 &  0.88 &  0.93 &  0.98 \\
0.9   &  0.92 &  0.94 &  0.97 &  0.99 \\
% 0.3   &  0.56 &  0.72 &  0.84 &  0.92 \\
% 0.5   &  0.67 &   0.80 &  0.89 &  0.94 \\
% 0.7   &  0.79 &  0.88 &  0.94 &  0.97 \\
% 0.9   &  0.92 &  0.95 &  0.97 &  0.99 \\
\bottomrule
\end{tabular}}\\
\vspace{7pt}
% m=400, n=100
\subfloat[$n=100$, $m=400$]{
% \begin{center}
\begin{tabular}{ccccr}
\toprule
\multirow{2}{*}{$\PP[\hat{Y}= Y\,;\, \Ycal]$} 
&\multicolumn{4}{c}{$\PP[Y' = Y]$}\\
& 0.3 & 0.5 & 0.7 &0.9 \\
\midrule
0.3   &  0.62 &  0.77 &  0.87 &  0.95 \\
0.5   &  0.73 &  0.83 &  0.91 &  0.97 \\
0.7   &  0.83 &  0.89 &  0.95 &  0.98 \\
0.9   &  0.93 &  0.96 &  0.98 &  0.99 \\
% 0.3   &  0.63 &  0.76 &  0.88 &  0.94 \\
% 0.5   &  0.73 &  0.84 &  0.92 &  0.96 \\
% 0.7   &  0.83 &   0.90 &  0.95 &  0.92 \\
% 0.9   &  0.93 &  0.96 &  0.98 &  0.96 \\
\bottomrule
\end{tabular}
% \end{center}
}
\caption{Empirical success probability achieved by four different experts using our system during test, each with a different success probability $\PP[\hat{Y}= Y\,;\, \Ycal]$, on four prediction tasks where the classifier achieves a different success probability $\PP[Y' = Y]$.
Each table corresponds to a different number of label values $n$ and calibration and estimation set size $m$.
For readability, each cell shows only the average since the standard errors are all below $10^{-2}$.} \label{table:synthetic-acc-additional}
\end{figure}

\clearpage
\newpage
\section{Beyond Standard Conformal Prediction}
\label{app:beyond-standard-cp}

In Section~\ref{sec:optimizing-conformal-prediction}, we have used standard conformal prediction~\citep{angelopoulos2021gentle} to construct the re\-commen\-ded subsets $\Ccal(X)$---we have constructed $\Ccal(X)$ by comparing the conformal scores $s(X, y)$ to a single thres\-hold $\hat{q}$, as shown in Eq.~\ref{eq:prediction-set}. 
%
% manuel: I do not think it is a good idea to write: "we will show that we can sometimes improve the performance of our system by",
% since we do not "show" it.
%
Here, we introduce a set-valued predictor based on conformal prediction that constructs $\Ccal(X)$ using two thresholds $\hat{q}_{\alpha_1}$ and $\hat{q}_{\alpha_2}$.
By doing so, the recommended subsets will include label va\-lues whose corresponding conformal 
scores are neither unreasona\-bly large, as in standard conformal prediction, nor unreasona\-bly low in comparison with the conformal scores of the samples in the calibration set $\Dcal_{\text{cal}}$.
This may be useful in scenarios where the classi\-fier underpinning our system has not particularly high ave\-rage accu\-ra\-cy\footnote{ In such scenarios, the conformal scores of the samples in the calibration set can occasionally have low values—otherwise, the classifier would be highly accurate---and thus it is beneficial to exclude label values with (very) low conformal scores from the recommended subsets---those label values the classifier is confidently wrong about.}.

More specifically, given a calibration set $\Dcal_{\text{cal}} = \{(x_i, s_i)\}_{i=1}^{m}$, let $\alpha_1, \alpha_2 \in [0,1]$, with $\alpha_1 < \alpha_2$, and $\hat{q}_{\alpha_1}$ and $\hat q_{\alpha_2}$ 
be the $\frac{\lceil(m + 1)(1 - \alpha_1)\rceil}{m}$ and $\frac{\lceil(m + 1)(1 - \alpha_2)\rceil}{m}$ empirical quantiles of the conformal scores $s(x_1, y_1), \ldots, s(x_m, y_m)$.
%
% Moreover, use the quantiles $\hat{q}_{\alpha_1}$ and $\hat{q}_{\alpha_2}$ to construct the subsets $\Ccal_{\alpha_1, \alpha_2}(X)$ for new data samples as follows:
 If we construct the subsets $\Ccal_{\alpha_1, \alpha_2}(X)$ for new data samples as follows:
\begin{equation} \label{eq:prediction-set-2}
    \Ccal_{\alpha_1,\alpha_2}(X) = \{y \given \hat q_{\alpha_2} < s(X, y) \leq \hat q_{\alpha_1} \},
\end{equation}
we have that the probability that the true label $Y$ belongs to the subset $\Ccal_{\alpha_1, \alpha_2}(X)$
conditionally on the calibration set $\Dcal_{\text{cal}}$ is almost exactly $\alpha_2 - \alpha_1$ with high probability as long as the size $m$ of the calibration set is sufficiently large. 
More specifically, we first note that the coverage probability is a random quantity whose distribution is given by the following proposition, which is the counterpart of 
Proposition~\ref{prop:coverage-beta}:
\begin{proposition}\label{prop:modified-cp-beta}
For a decision support system $\Ccal_{\alpha_1, \alpha_2}$ that constructs $\Ccal_{\alpha_1, \alpha_2}(X)$ using Eq.~\ref{eq:prediction-set-2}, as long as the conformal scores $s(x_i, y_i)$ for all $(x_i, y_i) \in \Dcal_{\text{cal}}$ are almost surely distinct, it holds that:
\begin{equation}\label{eq:modified-cp-beta}
    \PP[Y \in \Ccal_{\alpha_1, \alpha_2}(X) \given \Dcal_{\text{cal}}] \sim \text{Beta}(l, m - l + 1),
\end{equation}
where $l = \lceil(m+1)(1 - \alpha_1)\rceil - \lceil(m+1)(1 - \alpha_2)\rceil$.
%
% and the probability is over the randomness of the samples that the system $\Ccal_{\alpha_1, \alpha_2}$ helps predicting.
\end{proposition}
%
% Then, using the definition of the beta distribution, we have that
As an immediate consequence of Proposition~\ref{prop:modified-cp-beta}, using the definition of the beta distribution, we have that
\begin{equation*}
\alpha_2-\alpha_1 \leq \EE\left[ \PP[Y \in \Ccal_{\alpha_1, \alpha_2}(X) \given \Dcal_{\text{cal}}] \right]  = \alpha_2 - \alpha_1 + \frac{c_1 - c_2}{m+1} \leq \alpha_2 - \alpha_1 + \frac{1}{m+1},
\end{equation*}
where $c_1, c_2 \in [0, 1)$. 
Moreover, given a target probability $\alpha_2 - \alpha_1$ and tolerance values $\delta, \epsilon \in (0, 1)$, we can compute the minimum size $m$ of the calibration set $\Dcal_{\text{cal}}$
such that $\Ccal_{\alpha_1, \alpha_2}$ enjoys Probably Approximately Correct (PAC) coverage guarantees, \ie, with probability $1-\delta$, it holds that 
\begin{equation*}
\alpha_2 - \alpha_1 - \epsilon \leq \PP[Y \in \Ccal_{\alpha}(X) \given \Dcal_{\text{cal}}] \leq \alpha_2 - \alpha_1 + \epsilon.
\end{equation*}
Finally, given an estimator of the expert'{}s success pro\-ba\-bi\-lity $\hat{\mu}_{\alpha_1, \alpha_2}$ such that for each $\alpha_1 < \alpha_2$ and $\delta \in (0,1)$, with probability at least $1 - \delta$, it holds that $|\hat{\mu}_{\alpha_1,\alpha_2} - \PP[\hat Y = Y \,;\, \Ccal_{\alpha_1,\alpha_2}]| \leq \epsilon_{\alpha_1, \alpha_2, \delta}$,
we can proceed similarly as in standard conformal prediction to find the near optimal $\hat{\alpha}_1, \hat{\alpha}_2 \in \Acal$ that maximizes the expert'{}s success probability with high probability, by using $\hat{\mu}_{\alpha_1, \alpha_2}$ and $\epsilon_{\alpha_1, \alpha_2, 2\delta / (m(m-1))}$.
Here, it is worth pointing out that, in contrast with the case of standard conformal prediction, the time com\-ple\-xi\-ty of finding the near optimal $\hat{\alpha}_1$ and $\hat{\alpha}_2$ is $\Ocal(m \log m + mn \log n + m n^2)$. % because there are $m^2$ different pairs of $(\alpha_1, \alpha_2)$ values.
Moreover, we can still rely on the practical method to estimate the expert'{}s conditional success probability introduced in Section~\ref{sec:optimizing-conformal-prediction}.

\clearpage
\newpage

\section{Sensitivity to the Choice of Calibration Set} %Validity of the Conformal Guarantees}
\label{app:conf-guarantees}
In this section, we repeat the experiments on synthetic and real data using $100$ 
independent realizations of the calibration, estimation and test sets.
Then, for each data split, we compare the empirical coverage $\frac{1}{|\Dcal_{\text{test}}|} \sum_{(x, y) \in \Dcal_{\text{set}}} \II[ y \in C_{\hat{\alpha}}(x)] := 1 - \hat{\alpha}_{\text{emp}}$
achieved by our system $\Ccal_{\hat{\alpha}}$ on the test set $\Dcal_{\text{test}}$ to the
corresponding target coverage $1 - \hat{\alpha}$.

\begin{figure}[h]
    \centering
    \subfloat[Synthetic Task]{ \includegraphics[width=.49\columnwidth]{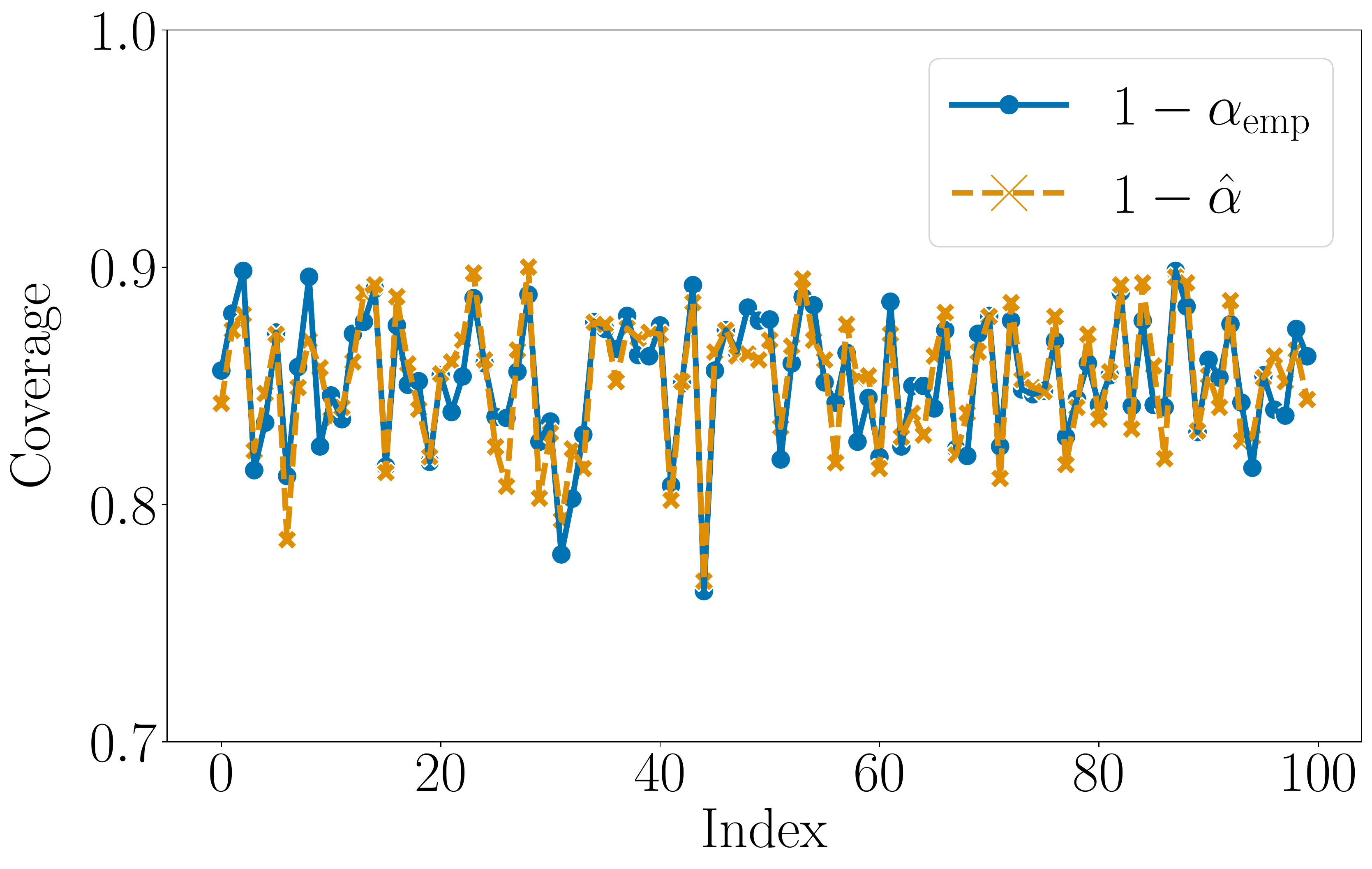}}
    % \vspace{-1mm}
    \subfloat[CIFAR-10H]{\includegraphics[width=.49\columnwidth]{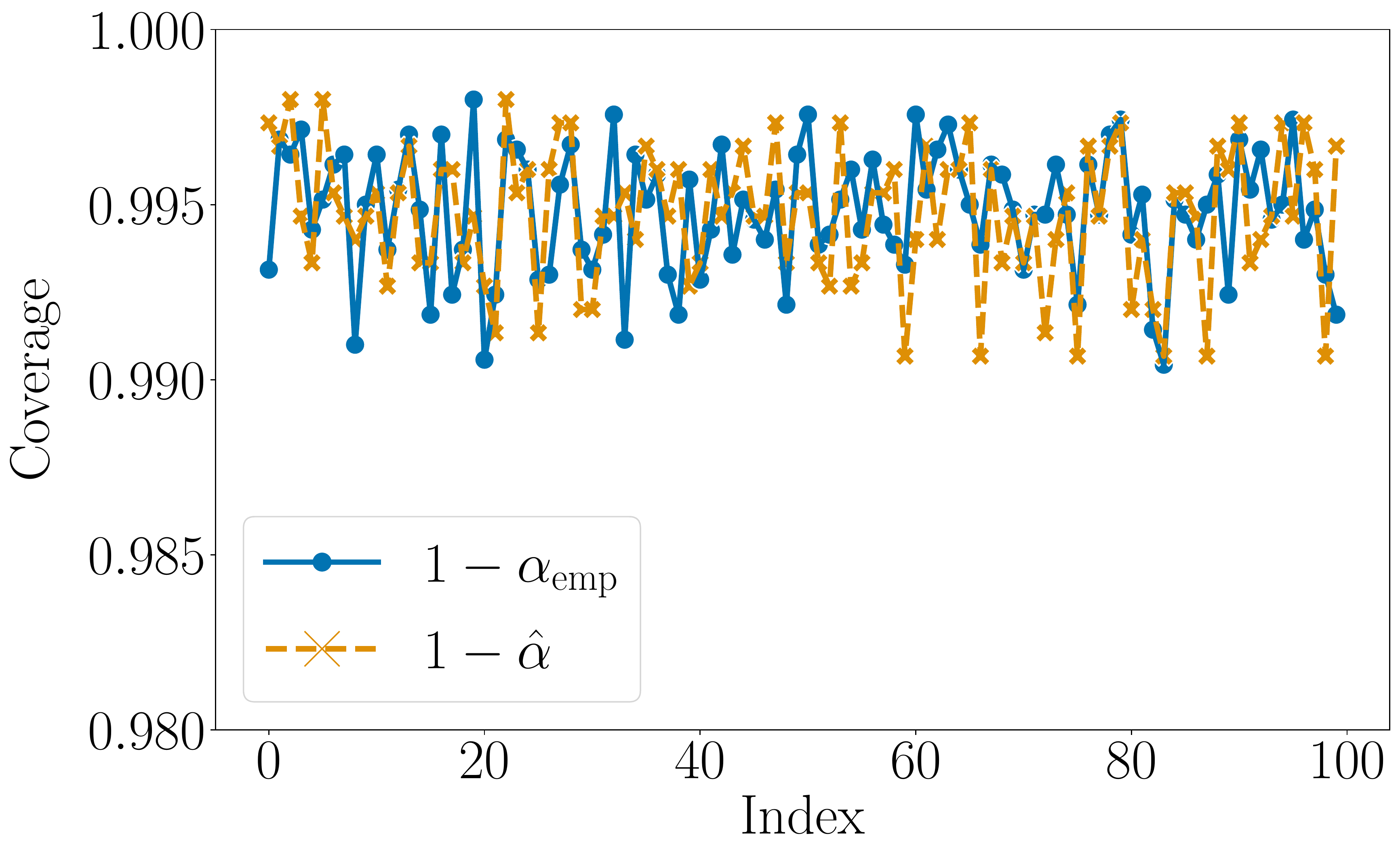}}
    \caption{Empirical test coverage $1 - \alpha_{\text{emp}}$ and target coverage 
    $1 - \hat{\alpha}$ for $100$ independent realizations of the calibration, estimation and test sets. 
    In Panel (a), the synthetic task comprises a classifier with $\PP[Y'=Y] = 0.5$ and an expert with $\PP[\hat{Y} = Y \,;\, \Ycal] = 0.5$, the number of labels is $n = 10$ and the size of the calibration and estimation sets is $m = 1{,}200$.
    In Panel (b), the classifier is the popular DenseNet classifier and $m=1{,}500$.}
    \label{fig:coverage}
\end{figure}

Figure~\ref{fig:coverage} summarizes the results for (a) one synthetic prediction task and one synthetic expert and (b) one popular deep neural network classifier on the CIFAR-10H dataset.
%
% we find the near optimal $\hat{alpha}$ using Algorithm~\ref{alg:near-optimal-alpha} and 
% measure the empirical coverage $1 - \alpha_{\text{emp}}$ during test, on both synthetic and 
% real data experiments. At test time, we used the same calibration set for which the near 
% optimal $\hat{\alpha}$ was computed. 
%
% We present the results for one synthetic prediction task and one expert, and on the dataset 
% CIFAR-10H with one popular deep neural network classifier in Figure~\ref{fig:coverage}. 
%
We find that the value of the near-optimal $\hat{\alpha}$ does not vary significantly 
across experiments (\ie, across calibration sets) and, for each experiment, the 
empirical coverage $1 - \hat{\alpha}_{\text{emp}}$ is very close to and typically higher than the target coverage $1 - \hat{\alpha}$.
We found similar results for other expert-classifier pairs with different success
probabilities.

% In both cases, not only do we observe that the near optimal $1 - \hat{\alpha}$ does not vary 
% significantly across the different calibration sets, but also that $1 - \hat{\alpha}$ is 
% always very similar to the empirical coverage $1 - \alpha_{\text{emp}}$ measured during test. 
% We also note that the mean $1 - \hat{\alpha}$  and $1 - \alpha_{\text{emp}}$ values are almost
% identical in both cases. The above observations indicate that for sufficiently large 
% calibration set, we can still have in practice approximate coverage guarantees, even if we fix
% the calibration set to compute the near optimal $\hat{\alpha}$.  

\clearpage
\newpage

\section{Success Probability Achieved by an Expert using Top-\texorpdfstring{$k$}{} Set-Valued Predictors}
\label{app:topk}
\begin{figure}[h!]
% \centering
\captionsetup[subfigure]{labelformat=empty,textfont=small}
 \foreach \human in {0.5, 0.9}{
        \foreach \machine in {0.5,0.7,0.9}{
            \subfloat[ $\PP{[\hat{Y}= Y\,;\, \Ycal]} = \human$, $\PP{[Y' = Y]} = \machine$]{
            \includegraphics[width=0.29\linewidth]{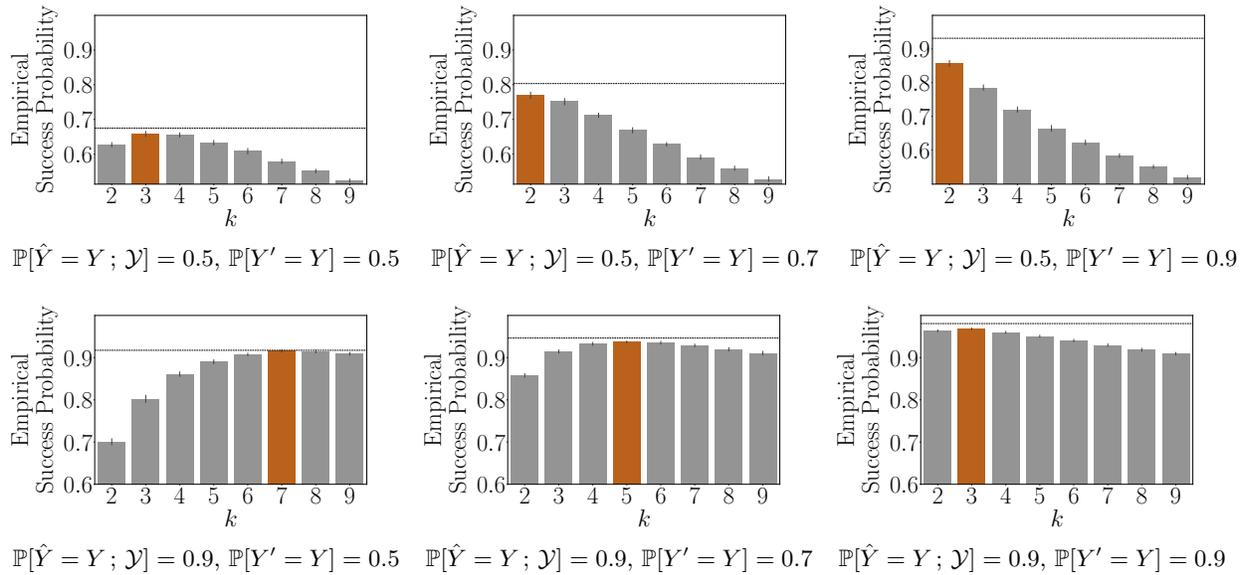} \quad}
        }}
\caption{Empirical success probability achieved by two different experts using the top-$k$ set-valued predictor ($\Ccal_{k}$) during test, each with a different success probability $\PP[\hat Y = Y \,;\, \Ycal]$, on three prediction tasks where the classifier achieves a different success probability $\PP[Y'=Y]$. 
In each panel, the horizontal dashed line shows the empirical success probability achieved by the same experts using our system ($\Ccal_{\hat{\alpha}}$) during test.
In all panels, the number of labels is $n = 10$, the size of the calibration and estimation sets is $m = 1{,}200$ and the results for the optimal $k$ value during test are highlighted in orange.}
  \label{fig:topk-synthetic}
\end{figure}

\begin{figure*}[!t]
% \centering
\captionsetup[subfigure]{labelformat=empty,textfont=small}
    \subfloat[\qquad ResNet-110, $\PP{[Y' = Y]} = 0.928$]{
    \includegraphics[width=0.3\columnwidth]{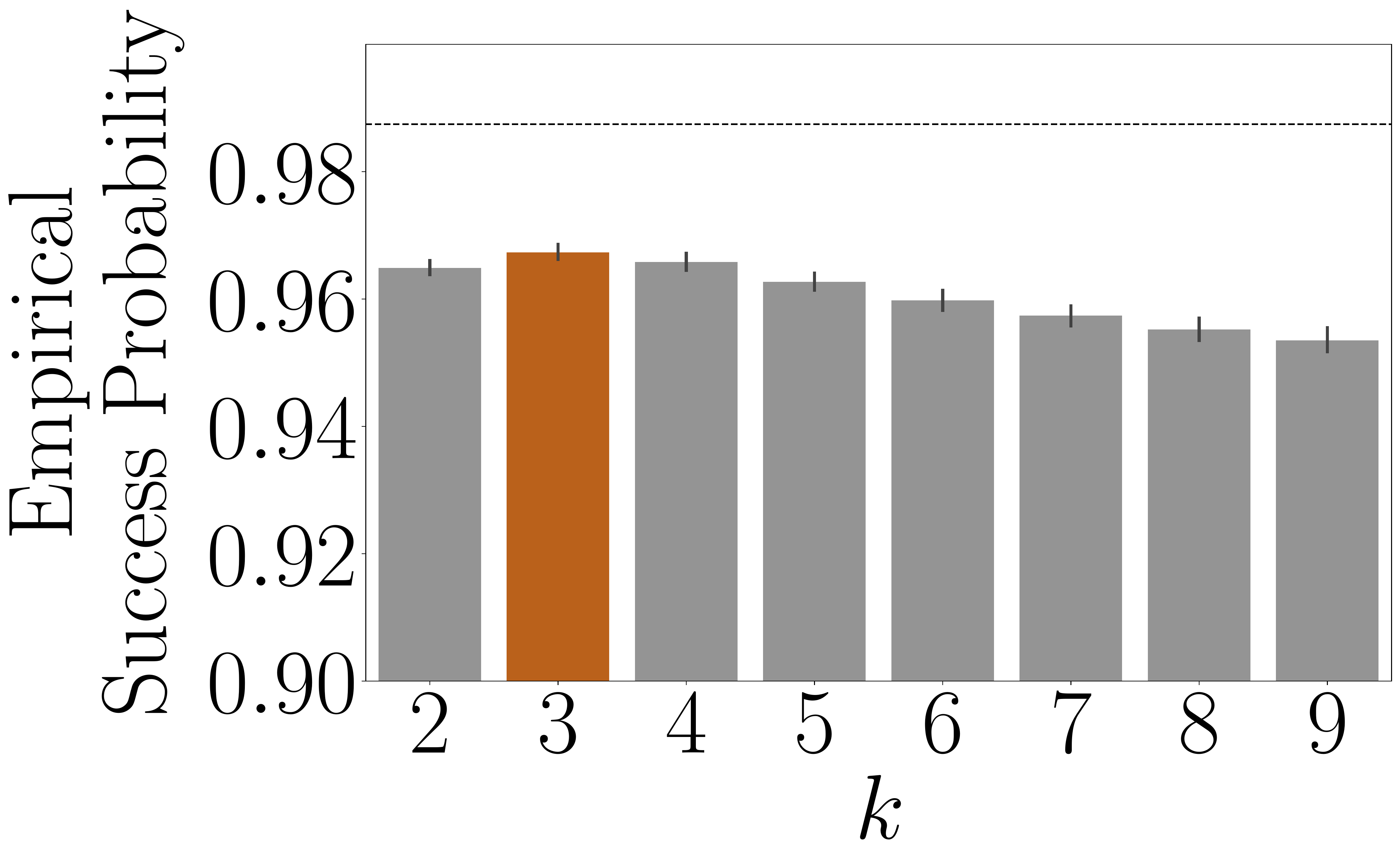} \quad}
    \subfloat[\qquad PreResNet-110, $\PP{[Y' = Y]} = 0.944$]{
    \includegraphics[width=0.3\columnwidth]{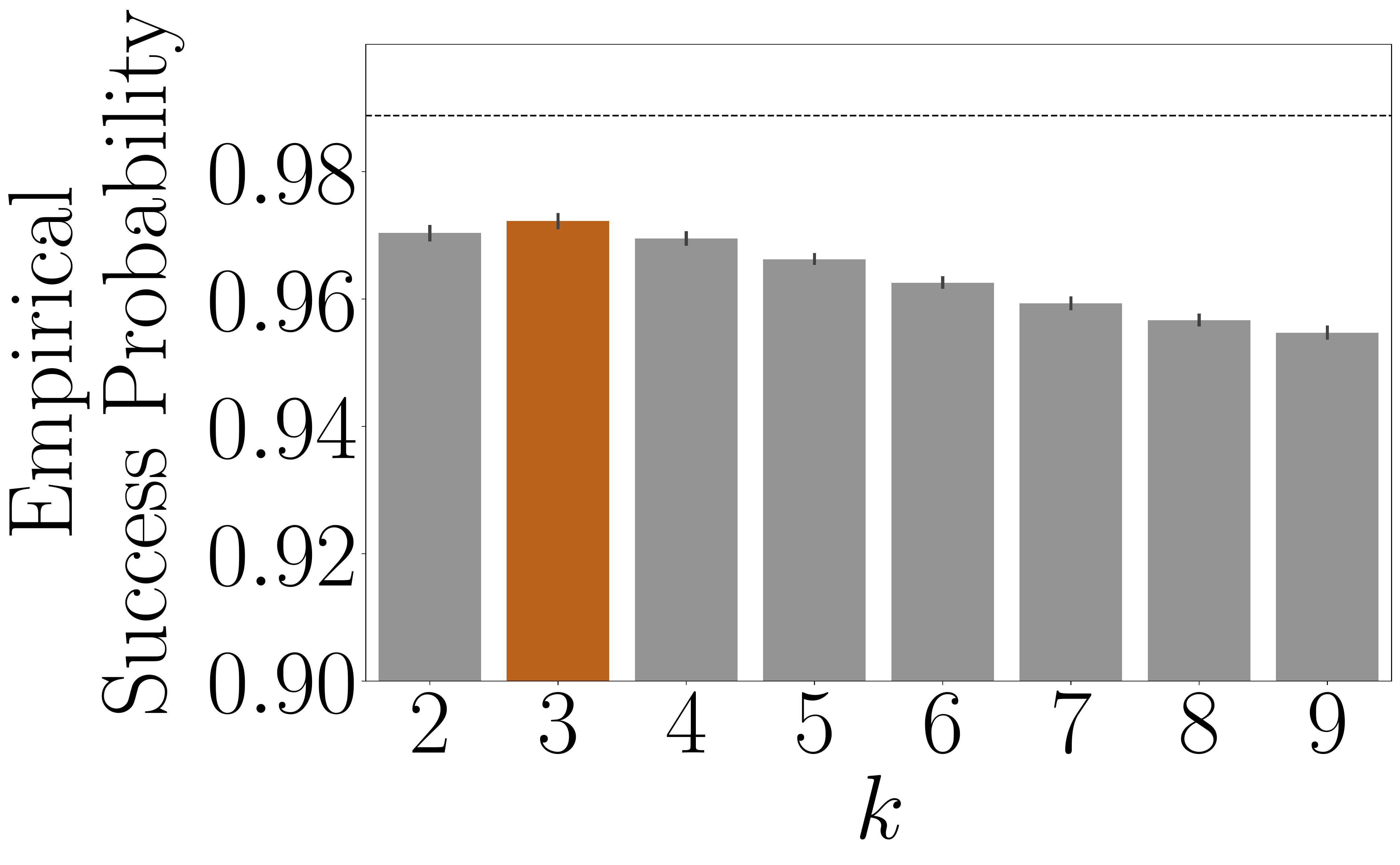} \quad }
    \subfloat[\qquad DenseNet, $\PP{[Y' = Y] =} 0.964$]{
    \includegraphics[width=.3\columnwidth]{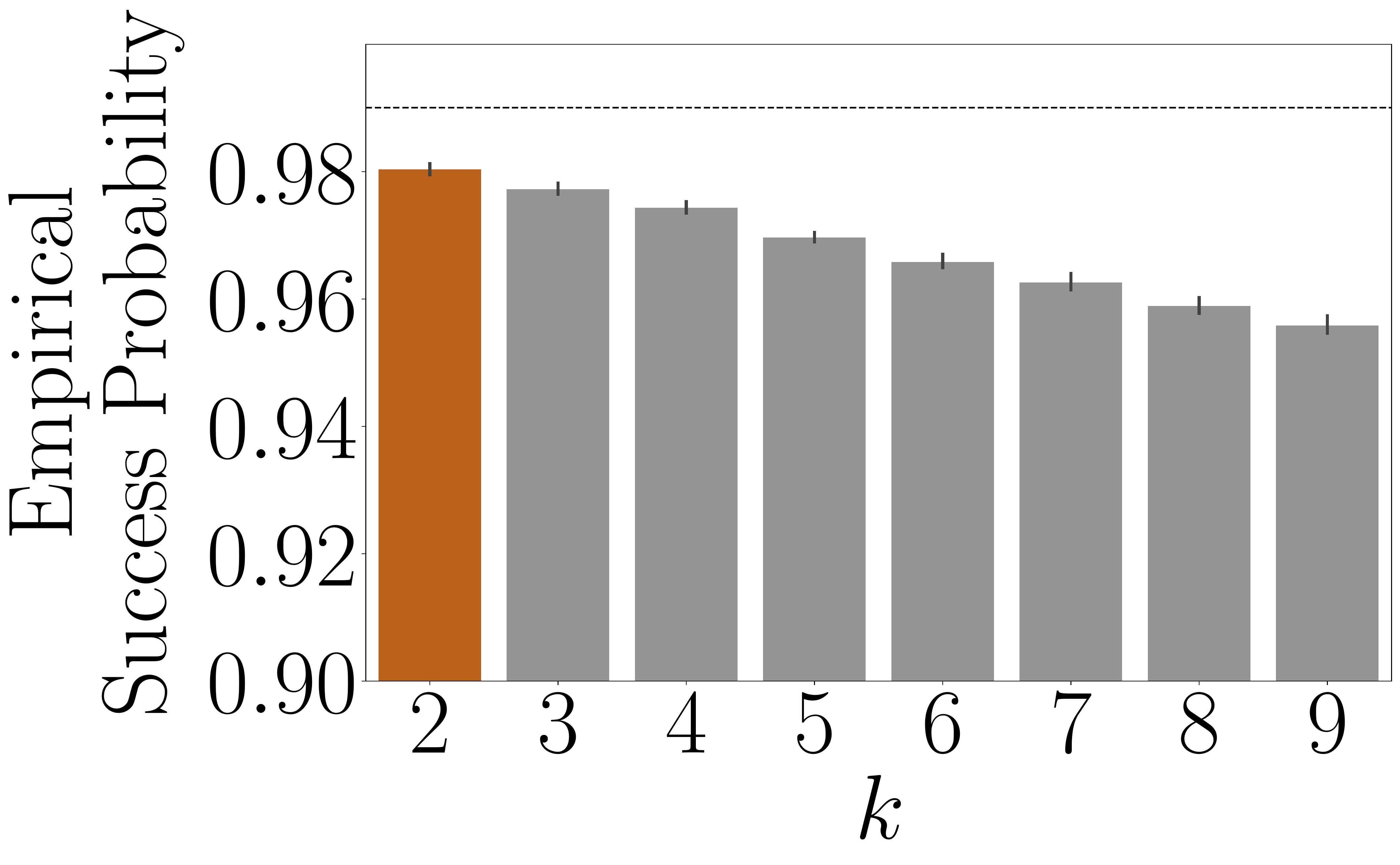}}
\caption{Empirical success probability achieved by an expert using three different top-$k$ predictors ($\Ccal_{k}$) during test, each with a different deep neural network classifier, on the CIFAR-10H dataset. 
In each panel, the horizontal dashed line shows an empirical success probability achieved by the same expert using our system ($\Ccal_{\hat{\alpha}}$) during test. 
In all panels, the size of the calibration and estimation sets is $m = 1{,}500$ and the results for the optimal $k$ value during test are highlighted in orange.}
% and the empirical success probability achieved by the expert at solving the 
% (original) multiclass task is $\PP[\hat{Y}= Y \,;\, \Ycal] = 0.947$.}
\vspace{150mm}
  \label{fig:topk-real}
\end{figure*}

In this section, we estimate the success probability achieved by an expert using the top-$k$ set-valued predictor for different $k$ values using both synthetic and real data.
Figures~\ref{fig:topk-synthetic} and~\ref{fig:topk-real} summarize the results, which show that, by allowing for recommended subsets of varying size, our system is consistently superior to the top-$k$ set-valued predictor across configurations.
Moreover, the results on synthetic data also show that, the higher the expert'{}s success probability $\PP[\hat Y = Y \,;\, \Ycal]$, the greater the optimal $k$ value (\ie, the greater the optimal size of the recommended subsets $\Ccal_{k}(X)$). 
This latter observation is consistent with the behavior exhibited by our system, where the higher the expert'{}s success probability $\PP[\hat Y = Y \,;\, \Ycal]$, the lower the value of the near-optimal $\hat \alpha$ and thus the greater the average size of the recommended subsets $\Ccal_{\hat \alpha}(X)$, as shown in Figure~\ref{fig:set-size-distr}.
%
% \vspace{-7mm}
% \newpage
% \clearpage
% \newpage

\clearpage
\newpage

\section{Size Distribution of the Recommended Subsets}
\label{app:set-size-distr}
\begin{figure}[h]
\centering
\captionsetup[subfloat]{labelformat=empty,textfont=small}
    \foreach \human in {0.5, 0.7, 0.9}{
        \foreach \machine in {0.5, 0.7, 0.9}{
            \subfloat[$\PP{[\hat{Y}= Y\,;\, \Ycal]} = \human$, $\PP{[Y' = Y]} = \machine$]{
            \includegraphics[width=0.32\columnwidth]{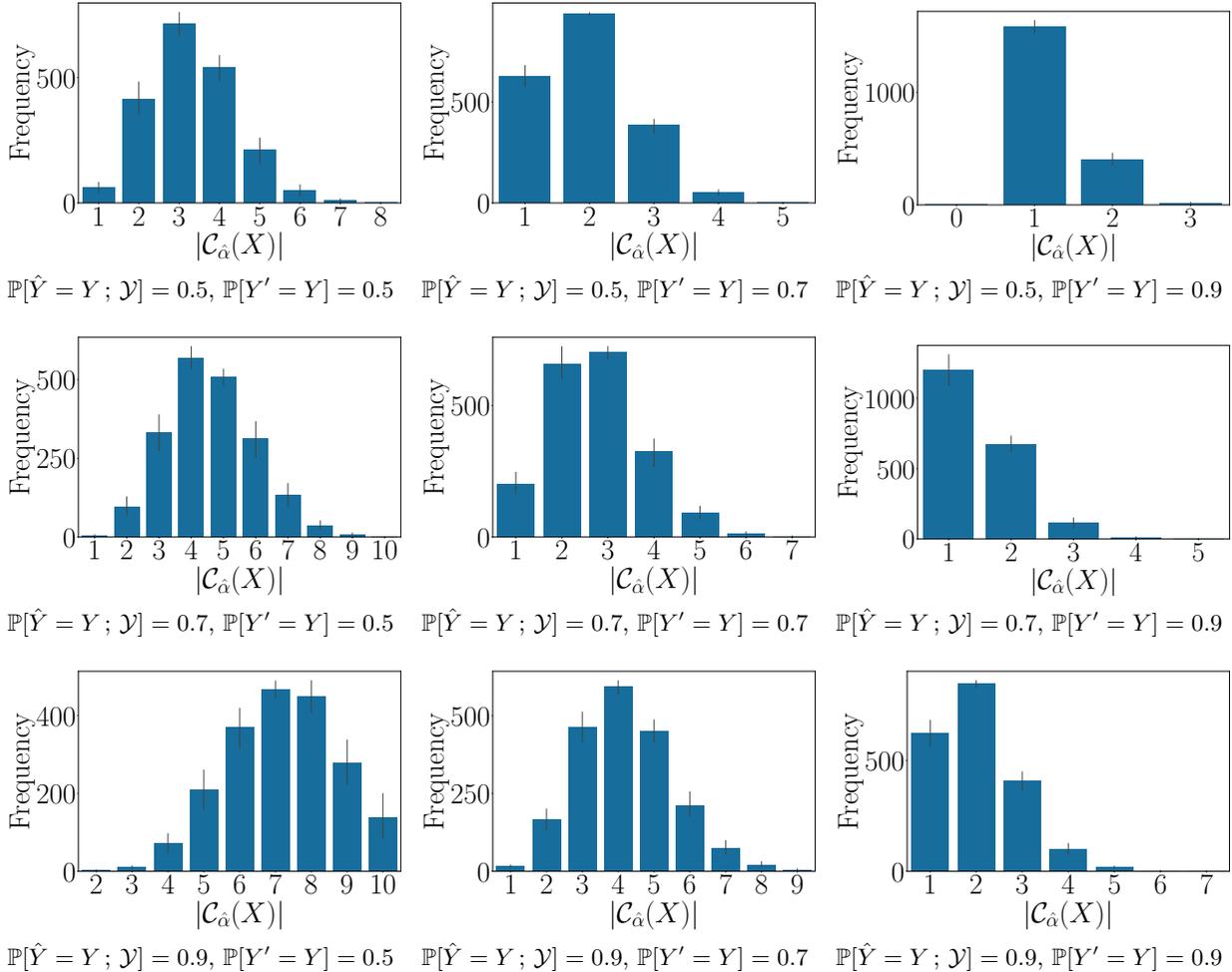}}
        }
    }
\caption{Empirical size distribution of the subsets $\Ccal_{\hat{\alpha}}(X)$ recommended by our system during test for different prediction tasks where the expert and the classifier achieve different success probabilities $\PP[\hat{Y} = Y \,;\, \Ycal]$ and $\PP[Y' = Y]$, respectively. In all panels, the number of labels is $n=10$ and the size of the calibration and estimation sets is $m=1{,}200$.}
  \label{fig:set-size-distr}
\end{figure}

Figure~\ref{fig:set-size-distr} shows the empirical size distribution of the subsets $\Ccal_{\hat{\alpha}}(X)$ recommended by our system during test for different experts 
and prediction tasks on synthetic data. 
The results show that, as the expert'{}s success probability $\PP[\hat Y = Y \,;\, \Ycal]$ increases and the near optimal $\hat \alpha$ decreases, the spread of the size distribution increases.

\clearpage
\newpage

\section{Performance of Our System Under Different \texorpdfstring{$\alpha$}{a} Values}
\label{app:alpha-error}

\begin{figure*}[h]
\centering
% \vspace{-3mm}
\subfloat[Empirical success probability vs $\alpha$]{\includegraphics[width=.5\columnwidth]{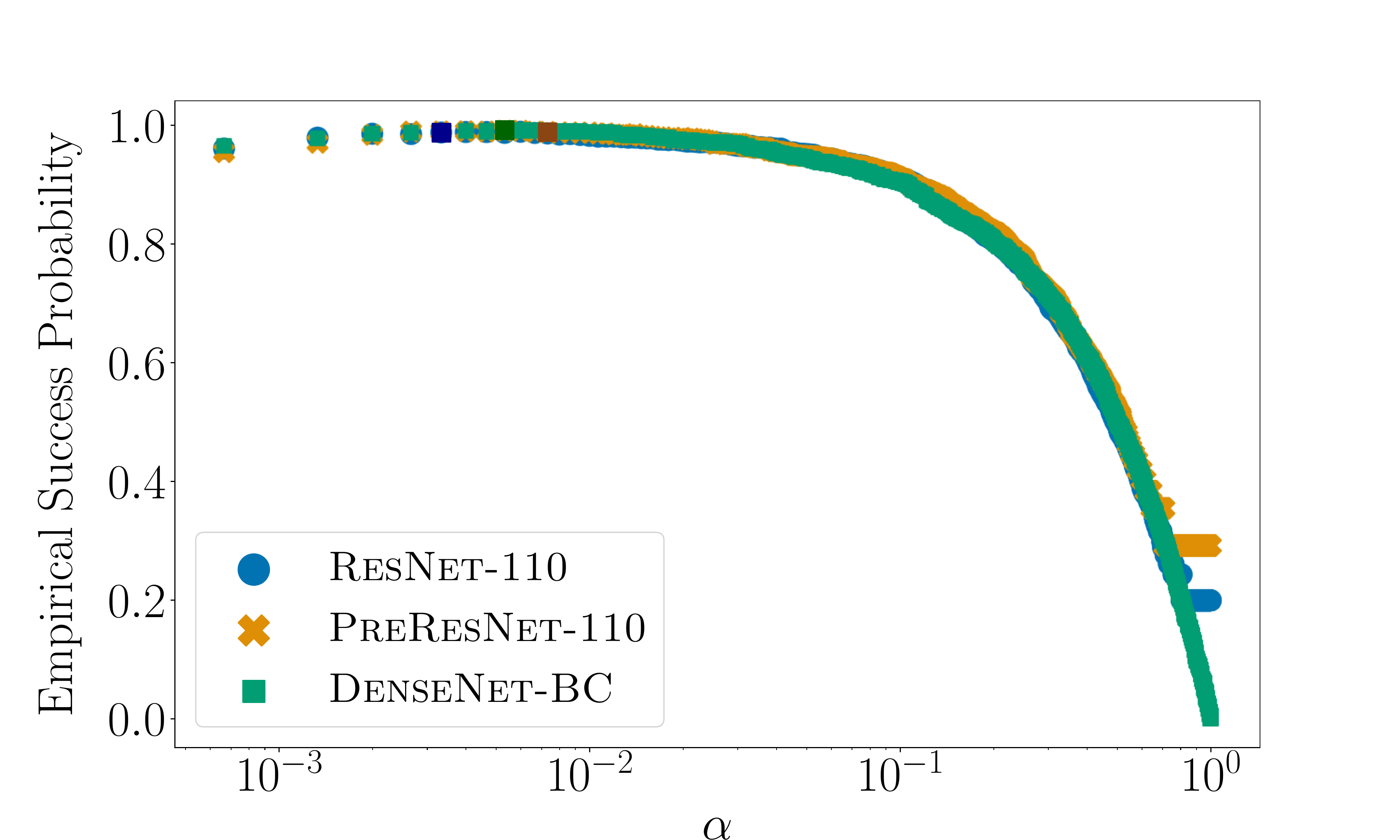}} 
\subfloat[Empirical average set size vs $\alpha$]{\includegraphics[width=.5\columnwidth]{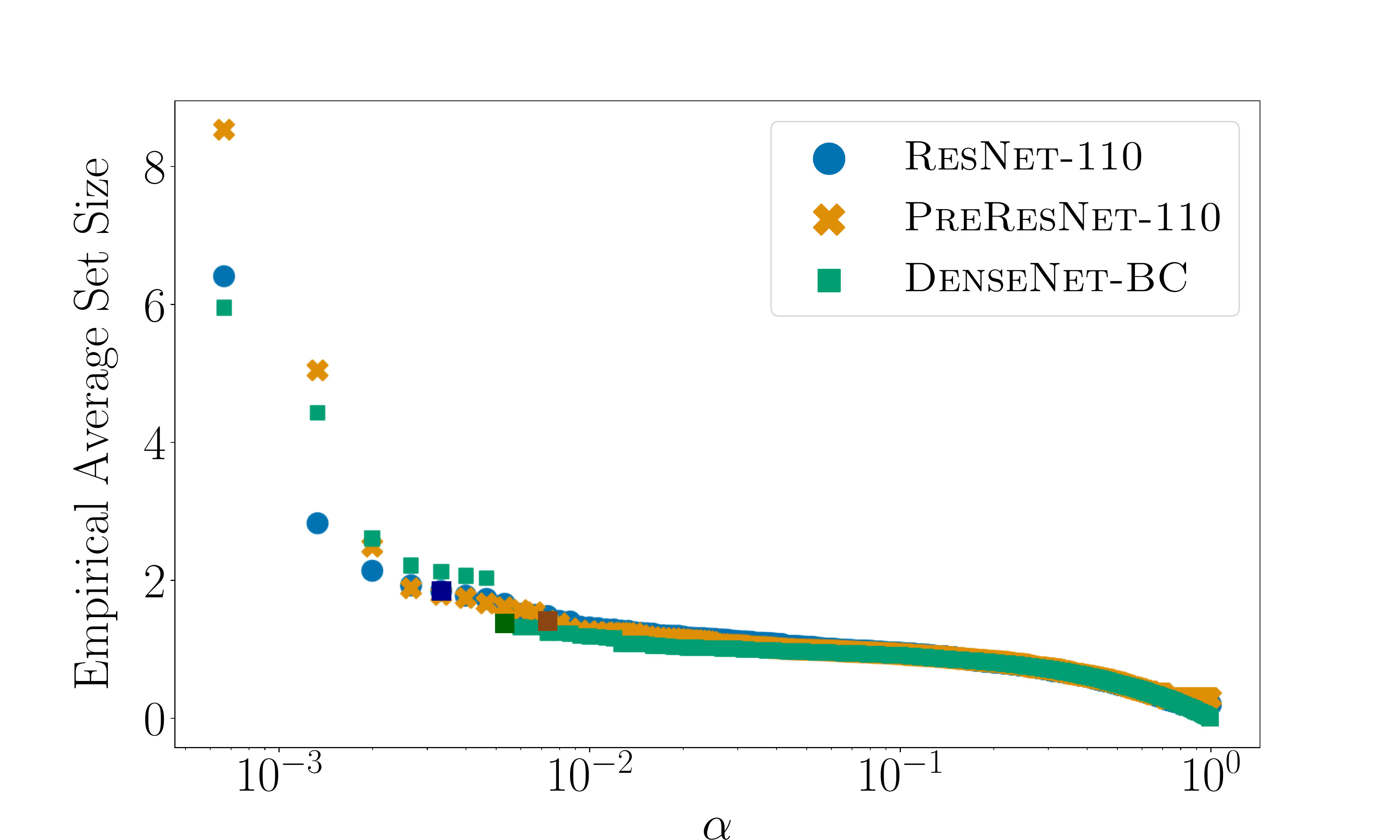}}
\caption{Empirical success probability achieved by an expert using our system
% $\PP[\hat{Y}= Y \,;\, \Ccal_{\alpha}]$
and average size of the recommended sets 
% $\EE{[|\Ccal_{\alpha}(X)|]}$ 
during test for each $\alpha \in \Acal$ and for three popular deep neural network classifiers on the CIFAR-10H dataset.
The size of the calibration and estimation sets is $m = 1{,}500$. 
Each marker corresponds to a different $\alpha$ value and the darker points correspond to $\hat{\alpha}$  for each task.}
\label{fig:real_error_alpha}
\vspace{-4mm}
\end{figure*}

In this section, we complement the results on CIFAR-10H dataset in the main paper by comparing, for each choice of classifier, the performance of our system under the near optimal $\hat{\alpha}$ found by Algorithm~\ref{alg:near-optimal-alpha} and under all other possible $\alpha$ values, inclu\-ding the optimal $\alpha^{*}$.
Figure~\ref{fig:real_error_alpha} summarizes the results, which suggest that, similarly as in the experiments on synthetic data, the performance of our system under $\hat{\alpha}$ and $\alpha^{*}$ is very similar. 
However, since the classifiers are all highly accurate, the average size of the recommended subsets under $\hat{\alpha}$ and $\alpha^{*}$ is quite close to one even though $\hat{\alpha}$ is much smaller than in the experiments in synthetic data.

\clearpage
\newpage

\section{Additional Experiments using an Estimator of the Expert'{}s Success Probability with a More Expressive Context}
\label{app:extra-expert-models}
In this section, we repeat the experiments on the CIFAR-10H dataset using 
an alternative discrete choice model with a more expressive context which, 
additionally to the true label, distinguishes between different levels of 
difficulty across data samples. 
The goal here is to show that our results are not an artifact of the 
choice of context used in the main paper.

We consider three increasing levels of difficulty, denoted as $L_{\text{easy}}$, $L_{\text{medium}}$, $L_{\text{hard}}$. 
The difficulty levels correspond to the 50\% and 25\% quantiles of the 
experts'{} fractions of correct predictions per sample in the (original) multiclass
classification task.
Samples with a fraction of correct predictions larger than the 50\% 
quantile belong to $L_{\text{easy}}$, those with a fraction of correct 
predictions smaller than the 25\% quantile belong to $L_{\text{hard}}$, 
and the remaining ones belong to $L_{\text{medium}}$.
Then, given a sample $(x,y)$ of difficulty $L$, we assume that the expert'{}s conditional success probability for the subset $\Ccal_{\alpha}(x)$ is given by:
\begin{equation}
    \label{eq:expert-conditional-more-expressive}
    \PP[\hat{Y} = y \,;\, \Ccal_{\alpha} \given y \in \Ccal_{\alpha}(x), L] = \frac{e^{u^{L}_{yy}}}{\sum_{y^\prime \in \Ccal_{\alpha}(x)}e^{u^{L}_{yy^\prime}}},
\end{equation}
where $u^{L}_{yy^\prime}$ denotes the expert preference for the label value $y^\prime \in \Ycal$ whenever the true label is $y$ and the difficulty level
of the sample is $L$. 

Further, to estimate the parameters $u^{L}_{yy^\prime}$, we resort to the conditional confusion matrix for the expert predictions on samples of 
difficulty $L$, \ie, 
$\mathbf{C}^L = \left[ C^{L}_{yy^{\prime}}\right]_{y, y^\prime \in \Ycal}\text{, where } C^{L}_{yy^{\prime}} = \PP[\hat{Y} = y^\prime \,;\, \Ycal \given Y = y, L]$,
and set $u^{L}_{yy^\prime} = \log C^{L}_{yy^{\prime}}$. 
Finally, we compute a Monte-Carlo estimate $\hat{\mu}_{\alpha}$ of the expert'{}s success probability $\PP[\hat{Y} = Y \,;\, \Ccal_{\alpha}]$ required by Algorithm~\ref{alg:near-optimal-alpha} using the above conditional success probability and an estimation set $\Dcal_{\text{est}} = \{(x_i, y_i)\}_{i \in [m]}$, \ie,
\begin{equation}
    \label{eq:estimator-more-expressive}
    \hat{\mu}_{\alpha} = \frac{1}{m}\sum_{i \in [m] \given y_i \in \Ccal_{\alpha}(x_i)} \PP[\hat{Y} = y_i \,;\, \Ccal_{\alpha}\given y_i \in \Ccal_{\alpha}(x_i), L(x_i)],
\end{equation}
where $L(x_i) \in \{L_{\text{easy}}, L_{\text{medium}}, L_{\text{hard}}\}$ denotes the difficulty level of $x_i$.

Table~\ref{table:real-acc-labels-more-expressive} summarizes the results, which suggest that, in agreement with the main paper, an expert using our system may solve the prediction task significantly more accurately than the expert or the classifier on their own. 
\begin{table}[ht]
 \caption{Empirical success probabilities 
% $\PP[Y' = Y]$ and $\PP[\hat Y = Y \,;\, \Ccal_{\hat{\alpha}}]$ 
achieved by three popular deep neural network classifiers and by an expert using our system with these classifiers during test on the CIFAR-10H dataset.
Here, we assume the expert follows the alternative discrete choice model defined by Eq.~\ref{eq:expert-conditional-more-expressive}.
The size of the calibration and estimation sets is $m = 1{,}500$ and the expert's empirical success probability at solving the (original) multiclass task is $\PP[\hat{Y}= Y \,;\, \Ycal] \approx 0.947$.
Each cell shows only the average since the standard errors are all 
below $10^{-2}$.} \label{table:real-acc-labels-more-expressive}
\begin{center}
% \begin{small}
    \small
    \begin{sc}
        \begin{tabular}{lcccr}
            \toprule
            \small
            &Classifier  & Expert using $\Ccal_{\hat{\alpha}}$\\
            \midrule
            ResNet-110    & 0.928 &  0.981  \\
            PreResNet-110    & 0.944 & 0.983  \\
            DenseNet    & 0.964 & 0.987  \\
            \bottomrule
        \end{tabular}
    \end{sc}
    % \end{small}
\end{center}
\vspace{-3mm}
\end{table}

% Similarly as in the main paper, we also find that the performance of the 
% system under $\hat{\alpha}$ is very similar to the performance under 
% $\alpha^{*}$ and the average size of the recommended subsets under 
% $\hat{\alpha}$ and $\alpha^{*}$ is close to one, as shown in 
% Figure~\ref{fig:real_alpha_more_expressive}.
%
Finally, similarly as in the main paper, we also found that our system performs well with a small amount of calibration and estimation data---the relative gain in empirical success probability achieved by an expert using our system with respect to the same expert on their own raises from $3.02\pm0.05\%$ under $m = 200$ to just $3.28\pm0.04\%$ under $m=1{,}500$.

\end{document}